\documentclass[lettersize,journal]{IEEEtran}
\input{paper.sty}
\usepackage{orcidlink}

\title{Task-Driven Detection of Distribution Shifts with Statistical Guarantees for Robot Learning}

\author{Alec Farid \orcidlink{0000-0003-4253-496X}, Sushant Veer \orcidlink{0000-0003-3498-6448}, Divyanshu Pachisia \orcidlink{0009-0007-5120-1899}, Anirudha Majumdar \orcidlink{0009-0002-2296-7485}
\thanks{
Alec Farid and Sushant Veer contributed equally to this work.

Alec Farid and Divyanshu Pachisia were with the Intelligent Robot Motion Lab, Princeton University, Princeton, NJ 08540 USA.

Sushant Veer was with the Intelligent Robot Motion Lab, Princeton University, Princeton, NJ 08540 USA. He is now with the Autonomous Vehicle Research Group, NVIDIA, Santa Clara, CA 95051 USA (e-mail: sveer@nvidia.com).

Anirudha Majumdar is with the Intelligent Robot Motion Lab, Princeton University, Princeton, NJ 08540 USA (e-mail: ani.majumdar@princeton.edu).
}}

\begin{document}
\maketitle
\begin{abstract}
Our goal is to perform \emph{out-of-distribution (OOD) detection}, i.e., to detect when a robot is operating in environments drawn from a different distribution than the ones used to train the robot. We leverage Probably Approximately Correct (PAC)-Bayes theory to train a policy with a \emph{guaranteed bound} on performance on the training distribution. Our idea for OOD detection relies on the following intuition: violation of the performance bound on test environments provides evidence that the robot is operating OOD. We formalize this via statistical techniques based on p-values and concentration inequalities. The approach provides guaranteed confidence bounds on OOD detection including bounds on both the false positive and false negative rates of the detector and is \emph{task-driven} and only sensitive to changes that impact the robot’s performance. We demonstrate our approach in simulation and hardware for a grasping task using objects with unfamiliar shapes or poses and a drone performing vision-based obstacle avoidance in environments with wind disturbances and varied obstacle densities. Our examples demonstrate that we can perform task-driven OOD detection within just a handful of trials. 
\end{abstract}

\section{Introduction}
\label{sec:introduction}

\begin{figure*}[t]
\begin{center}
\includegraphics[width=2\columnwidth]{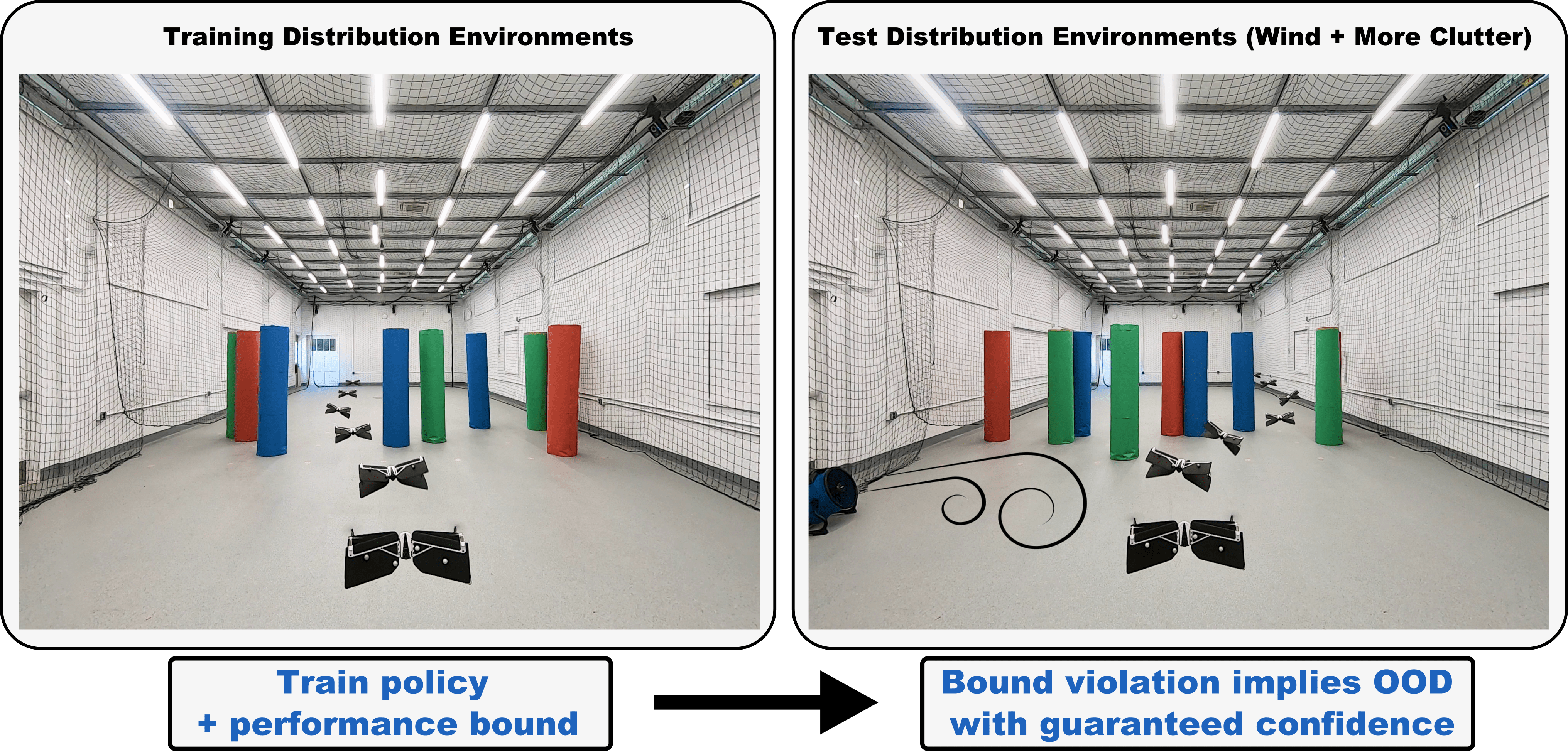}
\end{center} 
\caption{A schematic of our overall approach. We learn policies with guaranteed bounds on expected performance on the training distribution. Violation of this bound during deployment implies that the robot is operating OOD (with high confidence). We present hardware experiments for a drone navigating in new environments with varying wind conditions and clutter, along with simulation and hardware experiments for a grasping task. \label{fig:anchor}}
\vspace{-12pt}
\end{figure*}

Imagine a drone trained to perform vision-based navigation using a dataset of indoor environments and deployed in environments with varying wind conditions or obstacle densities (Figure~\ref{fig:anchor}). Similarly, consider a robot arm manipulating a new set of objects or an autonomous vehicle deployed in a new city. State-of-the-art techniques for learning-based control of robots typically struggle to generalize to such \emph{out-of-distribution} (OOD) environments. This lack of OOD generalization is particularly pressing in safety-critical settings, where the price of failure is high. In this work, we focus on the problem of autonomously \emph{detecting} when a robot is operating in environments drawn from a different distribution than the one used to train the robot. This ability to perform \emph{OOD detection} has the potential to improve the safety of robotic systems operating in OOD environments. For example, a drone operating in a new set of environments could either deploy a highly conservative policy or cease its operations altogether. In addition, OOD detection can also allow the robot to improve its policy by re-training using additional data collected from the new environments. 

There are two important desiderata that OOD detection approaches for safety-critical robotic systems should ideally satisfy. First, we would like to develop OOD detection techniques with \emph{guaranteed confidence bounds}. Second, we would like our OOD detectors to be \emph{task-driven} and only sensitive to \emph{task-relevant} changes in the robot’s environment. As an example, consider again the drone navigation setting in Figure~\ref{fig:anchor} and suppose that the robot’s policy is insensitive to changes in color and lighting. Here, the robot's OOD detector should \emph{not} trigger even if the robot is operating in environments with different color/lighting and should only trigger if there are task-relevant variations (e.g., variations in the obstacle density). Unfortunately, current approaches (Section~\ref{sec:related work}) do not typically satisfy both desiderata; they are often based on heuristics and are not task-driven in general. 

\emph{Statement of Contributions.} We develop task-driven OOD detection techniques with statistical guarantees on correctness.  To this end, we make four specific contributions (see Figure~\ref{fig:anchor} for an overview).
\begin{itemize}[leftmargin=*]
\item Given a dataset of environments drawn from an (unknown) training distribution, we develop a pipeline based on \emph{generalization theory} for training control policies with a \emph{guaranteed bound} on performance (a bound on the expected cost of the policy on the unknown training distribution). Specifically, we leverage recently developed \emph{ derandomized probabilistically correct (PAC)-Bayes} bounds that are well suited to enable OOD detection (Section~\ref{subsec:PAC-Bayes}). 

\item We develop two OOD detection techniques (Section~\ref{subsec:ood-detect-theory}), using p-values and concentration inequalities, with complementary statistical interpretations. Both detectors are based on the following intuition: if the costs incurred when the robot is deployed in a small number of new environments violate the bound on the policy's performance, this indicates that the robot is operating OOD. Since our OOD detection scheme leverages the costs incurred in new environments, it is \emph{only} triggered by task-relevant changes. In particular, we identify two distinct OOD events: OOD-adverse~($\oodRightSym$) and OOD-benign~($\oodLeftSym$), which correspond to OOD events that result in costs that are higher than the PAC-Bayes generalization bound and lower than the PAC-Bayes generalization bound, respectively. Both $\oodRightSym$ and $\oodLeftSym$ are OOD events, but the former is detrimental to the robot (requiring an intervention) while the latter is not.


\item Our detection schemes have the ability to perform OOD detection with guaranteed confidence bounds. This allows us to provide statistical guarantees on \emph{both} the false positive rate (probability that $\oodRightSym$ is incorrectly detected) and the false negative rate (probability that $\oodLeftSym$ is incorrectly detected) for our detectors; positive detection is one that requires intervention to the robot's nominal operation whereas a negative detection is one that does not.

\item We demonstrate our approach on two simulated examples (Section~\ref{sec:examples}): (i) a robotic manipulator grasping a new set of objects in varying locations, and (ii) a drone navigating a new set of environments. Comparisons with baselines demonstrate the advantages of our approach in terms of providing statistical guarantees and being insensitive to task-irrelevant shifts. We also present a thorough set of hardware experiments for vision-based drone navigation with varying wind conditions and clutter (Figure~\ref{fig:anchor}) and for grasping with varying objects and poses. Our experiments demonstrate the ability of our approach to perform task-driven OOD detection within just a handful of trials for systems with complex dynamics and rich sensing modalities. 
\end{itemize} 

A preliminary version of this work was presented in the Conference of Robot Learning (CoRL) 2021 \citep{Farid21}. In this significantly extended and revised version, we additionally present: (i) an extension of our OOD detection methods to also detect OOD-benign~($\oodLeftSym$) environments (Section~\ref{subsec:ood-detect-theory}), (ii) formulations of our detectors in terms of algorithms that output the detectors' predictions (Algorithm~\ref{alg:HT-detector} and Algorithm~\ref{alg:CI-detector}), (iii) bounds on the false negative rate on the confidence-interval based OOD detector (Remark~\ref{rem:fp_fn_guarantee}), (iv) hardware results on the OOD detector for the grasping example previously studied in simulation (Section~\ref{subsec:grasping}) and (v) expanded simulation results and a study of the effect of the cost function and chosen confidence bounds for the navigation example (Section~\ref{subsec:swing}).
\section{Related work}
\label{sec:related work}

\textbf{Anomaly/OOD detection in supervised learning.} 
Anomaly detection in low-dimensional signals has been well-studied in the signal processing literature (see \cite{Basseville88} for a review). Recent work in machine learning has focused on OOD detection for high-dimensional inputs (e.g., images) in supervised learning settings (see \cite{Ruff21} for a review). Popular approaches use threshold-based detectors for the output distribution of a given pre-trained neural classifier \citep{Hendrycks17, Liang18, Hendrycks20}. Other methods use a specific training pipeline in order to improve OOD detection on test samples \citep{DeVries18, Lee18b, Hendrycks19, Zisselman20}. However, these methods are often susceptible to adversarial attacks \citep{Chen20}. Thus, approaches for addressing adversarial data have been developed \citep{Lee18a, Chen20, Bitterwolf20}. Some of these approaches are also able to provide theoretical guarantees of performance on adversarial data \citep{Chen20, Meinke20, Meinke21}. Other methods provide PAC-style statistical guarantees \citep{Siddiqui16, Liu18, Fang22} or p-values \citep{Bates21}. However, these methods typically focus on supervised learning settings and often require specific network outputs (e.g., softmax) that are incompatible with non-classification tasks. In contrast, we focus on OOD detection for policy learning settings in robotics and do not make assumptions about the specific structure of the policy.

{\bf Task-driven OOD detection.} The methods above are aimed at detecting \emph{any} distributional shift in the data and can be sensitive even to task-\emph{irrelevant} shifts (i.e., ones that do not impact performance) as we demonstrate in our experiments (Section~\ref{sec:examples}). 
A recent method determines an estimate of input atypicality for pre-trained networks and uses it as an OOD detector in supervised learning settings \citep{Sharma21}. Another approach performs novelty detection on images from a vision-based robot for collision avoidance \citep{Richter17}. 
Recent methods have also been developed specifically for reinforcement learning (RL) \citep{Sedlmeier19, Sedlmeier20, Cai20, Wu21, Greenberg21}. In particular, \cite{Greenberg21} presents a general task-driven approach for OOD detection on sequential rewards, which is optimal in certain settings. 
However, neither this method nor others in the RL context provide statistical guarantees on detection. We propose an OOD detection framework which is both task-driven and provides statistical guarantees by leveraging generalization theory.

\textbf{Generalization theory.} 
Generalization theory provides a way to learn hypotheses (in supervised learning) with a bound on the true expected loss on the underlying data-generating distribution given only a finite number of training examples. Original frameworks include Vapnik-Chervonenkis (VC) theory \citep{Vapnik68} and Rademacher complexity \citep{Shalev14}. However, these methods often provide vacuous generalization bounds for high-dimensional hypothesis spaces (e.g., neural networks). Bounds based on PAC-Bayes generalization theory \citep{Shawe-Taylor97, McAllester99, Seeger02} have recently been shown to provide strong guarantees in a variety of settings \citep{Dziugiate17, Langford03, Germain09, Bartlett17, Jiang20, Perez-Ortiz20,Lotfi22}, and have been significantly extended and improved \citep{Catoni04, Catoni07, McAllester13, Rivasplata19, Thiemann17, Dziugaite18}. PAC-Bayes has also recently been extended to learn policies for robots with guarantees on generalization to novel environments \citep{Majumdar18, Veer20, Ren20, Majumdar21}. In the present work, we leverage recently-proposed \emph{derandomized} PAC-Bayes bounds \citep{Viallard21}; this framework allows us to train a single deterministic policy with a guaranteed bound on expected performance on the training distribution (in contrast to \cite{Majumdar18, Veer20, Ren20, Majumdar21}, which train stochastic neural network policies). This forms the basis for our OOD detection framework: by observing violations of the PAC-Bayes bound on test environments, we are able to perform task-driven OOD detection with statistical guarantees.

\section{Problem formulation}
\label{sec:problem_formulation}

{\bf Dynamics and environments.} Let $s_{t+1} = f_E(s_t, a_t)$ describe the robot's dynamics, where $s_t \in \mathcal{S} \subseteq \RR^{n_s}$ is the state of the robot at time-step $t$, $a_t \in \mathcal{A} \subseteq \RR^{n_a}$ is the action, and $E \in \mathcal{E}$ is the environment that the robot is operating in. ``Environment" here broadly refers to factors that are external to the robot, e.g., a cluttered room that a drone is navigating, disturbances such as wind gusts, or an object that a manipulator is grasping. The dynamics of the robot may be nonlinear/hybrid. We denote the robot's sensor observations (e.g., RGB-D images) by $o_t \in \mathcal{O} \subseteq \RR^{n_o}$. 

{\bf Cost functions.} The robot's task is encoded via a cost function and we let $C_E(\pi)$ denote the cost incurred by a (deterministic) policy $\pi$ when deployed in environment $E$ over a finite time horizon $T$. The policy $\pi \in \Pi$ is a mapping from (histories of) sensor observations to actions (e.g., parameterized using a neural network). In the context of obstacle avoidance, the cost could capture how close the drone gets to an obstacle; in the context of grasping, the cost could be 0 if the robot successfully lifts the object and 1 otherwise. We assume that the cost is bounded; without further loss of generality, we assume $C_E(\pi) \in [0,1]$. We also assume that the robot has access to the cost $C_E(\pi)$ after performing a rollout on $E$ (i.e., at the end of an episode of length $T$). This is a relatively benign assumption in robotics contexts since the cost often has physical meaning and can be measured by the robot's sensors. For example, a drone equipped with a depth sensor can measure the smallest reported depth value during its operation in an environment, and a manipulator equipped with a camera or
force-torque sensor can measure if it successfully grasped an object. We make no further assumptions on the cost function (e.g., we \emph{do not} assume continuity or Lipschitzness). 

{\bf Training and testing distribution.} We assume that the robot has access to a training dataset $S = \{E_1, \dots, E_m \}$ of $m$ environments drawn i.i.d. from a training distribution $\D$, i.e. $S \sim \D^m$. After training, the robot is deployed on environments in $S' = \{E'_1, \dots, E'_{n}\}$ drawn from a test distribution $\D'$: $S' \sim \D'^n$. Importantly, we \emph{do not} assume any explicit knowledge of $\D, \D'$, or the space $\mathcal{E}$ of environments. We only have indirect access to $\D$ and $\D'$ in the form of the finite training datasets $S$ and $S'$.

{\bf Goal: task-driven OOD detection with statistical guarantees}. After being deployed in (a typically small number of) environments in $S'$, the robot's goal is to detect if these environments were drawn from a different distribution than the training distribution (i.e., if $\D'$ is different from $\D$). Moreover, our goal is to perform \emph{task-driven} OOD detection. In particular, we consider environments drawn from $\D'$ as OOD-adverse~if $\D'$ satisfies the following: 
\begin{equation}
\label{eq:task-driven OOD}
 \ C_{\D'}(\pi) \coloneqq \underset{E' \sim \D'}{\EE} C_{E'}(\pi) > C_{\D}(\pi) \coloneqq \underset{E \sim \D}{\EE} C_{E}(\pi) ,
\end{equation}
and OOD-benign~if
\begin{equation}
\label{eq:task-driven WD}
 \ C_{\D'}(\pi) \leq C_{\D}(\pi) .
\end{equation}
Thus, our OOD-adverse detector should be \emph{insensitive} to changes in the environment distribution that do not adversely impact the robot's performance. This is a challenging task since we only assume access to a finite number of environments from $\D$ and $\D'$. Moreover, our goal is to develop an OOD detection framework that is broadly applicable in challenging settings involving nonlinear/hybrid dynamics, rich sensing modalities (e.g., RGB-D), and neural network-based policies.

\section{Approach}
\label{sec:Approach}

Our overall approach is illustrated in Figure~\ref{fig:anchor}. First, we train a policy with an associated \emph{guarantee} on the expected cost on the training distribution $\D$ (Section~\ref{subsec:PAC-Bayes}). We then apply our OOD detection scheme which formalizes the following intuition: violation of the bound during deployment implies (with high confidence) that the test distribution $\D'$ is OOD in a task-relevant manner (Section~\ref{subsec:ood-detect-theory}).

\subsection{Policy training via derandomized PAC-Bayes bounds}
\label{subsec:PAC-Bayes}

Given a training dataset $S = \{E_1, \dots, E_m \}$ of $m$ environments drawn i.i.d. from the training distribution $\D$, our goal is to learn a policy $\pi$ with a guaranteed bound on the expected cost $C_\D(\pi) \coloneqq \EE_{E \sim \D} C_E(\pi)$. Since our OOD detection scheme will rely on violations of the bound, it is important to obtain bounds that are as tight as possible. In this work, we utilize the \emph{Probably Approximately Correct (PAC)-Bayes} framework \citep{Shawe-Taylor97, McAllester99, Seeger02} to train policies with strong guarantees. More specifically, we leverage recently developed \emph{ derandomized} PAC-Bayes bounds \citep{Viallard21}, which are well-suited to the OOD detection setting (as we explain further below).

PAC-Bayes applies to settings where one chooses a \emph{distribution} over policies (e.g., a distribution over weights of a neural network), and learning algorithms that have the following structure: (1) choose a \emph{``prior"} distribution $P_0$ over the policy space $\Pi$ \emph{before} observing any data (this can be used to encode domain/expert knowledge); (2) obtain a training dataset $S$ and choose a \emph{posterior} distribution $P$ over the policy space $\Pi$. Let $P$ be the output of an algorithm $A$ which takes $P_0$ and $S$ as input. Denote the cost incurred by a policy $\pi$ on the training environments in $S$ as $C_S(\pi) := \frac{1}{m} \sum_{E \in S} C_E(\pi)$. 
The following result is our primary theoretical tool for training policies with bounds on performance.
\begin{theorem}
\label{thm:derand_pacbayes_specialized}
For any distribution $\D$, prior distribution $P_0$, $\delta \in (0,1)$, cost bounded in $[0,1]$, $m\geq 8$, and deterministic algorithm $A$ which outputs the posterior distribution $P$, we have the following:
\begin{align}
    \label{pointwise_pacbayes_specialized}
    \underset{(S, \pi) \sim (\D^m \times P)}{\PP}\ \bigg[ C_\D(\pi) \leq \overline{C}_\delta(\pi,S)\bigg] \geq 1 - \delta ,
\end{align}
 where $\overline{C}_\delta(\pi,S):= C_S(\pi) + \sqrt{R}$, $R:=\big(D_2(P \| P_0) + \ln\frac{2\sqrt{m}}{(\delta/2)^3}\big)/(2m)$, and $D_2$ is the R\'enyi Divergence for $\alpha = 2$ defined as: $    D_2(P||P_0) = \ln \big(\EE_{\pi \sim P_0}\big[\big(\frac{P(\pi)}{P_0(\pi)}\big)^2\big]\big)$.
\begin{proof} 
The proof is in Appendix~\ref{ap:proof derand pacbayes}. We use \citep[Theorem 2]{Viallard21}, a general pointwise PAC-Bayes bound. We perform the reduction from supervised learning to policy learning presented in \citep{Majumdar18}.
\end{proof}
\end{theorem} \vspace{-7pt}
We can provide a lower bound on $C_\mathcal{D}$ as an immediate corollary of the above theorem.
\begin{corollary}\label{cor:pac-bayes-lower-bound}
Let the assumptions of Theorem~\ref{thm:derand_pacbayes_specialized} hold. Then
\begin{align}
    \label{pac-bayes-lower}
    \underset{(S, \pi) \sim (\D^m \times P)}{\PP}\ \bigg[ C_\D(\pi) \geq \underline{C}_\delta(\pi,S)\bigg] \geq 1 - \delta ,
\end{align}
where $\underline{C}_\delta(\pi,S):= C_S(\pi) - \sqrt{R}$.
\end{corollary}
\begin{proof}
Note that Theorem~\ref{thm:derand_pacbayes_specialized} holds for any cost function bounded between $[0,1]$. Define a function $\hat{C}_E(\pi) := 1 - C_E(\pi)$. Analogously, we also have $\hat{C}_S(\pi) = 1 - C_S(\pi)$ and $\hat{C}_\mathcal{D}(\pi) = 1 - C_\mathcal{D}(\pi)$. Furthermore, $C_E(\pi)\in[0,1] \iff 1 - C_E(\pi)\in[0,1] \iff \hat{C}_E(\pi)\in[0,1]$. Hence, we can apply Theorem~\ref{thm:derand_pacbayes_specialized} on $\hat{C}$ to obtain:
\begin{align} \nonumber
    \underset{(S, \pi) \sim (\D^m \times P)}{\PP}\ & \bigg[ \hat{C}_\D(\pi) \leq \hat{C}_S(\pi) + \sqrt{R}\bigg] \geq 1 - \delta \\ 
    \nonumber \implies \underset{(S, \pi) \sim (\D^m \times P)}{\PP}\ & \bigg[C_\D(\pi) \geq C_S(\pi) - \sqrt{R}\bigg] \geq 1 - \delta, 
\end{align}
completing the proof.
\end{proof}
These results allow us to obtain policies with guaranteed upper and lower bounds on the expected cost. In particular, we can search for a posterior $P$ in order to minimize the upper bound $\overline{C}_\delta(\pi,S)$, i.e., in order to minimize the sum of the training cost and the ``regularizer" $\sqrt{R}$. We describe such training methods via backpropagation and blackbox optimization in Appendix~\ref{ap:backprop alg} and \ref{app:reg-NES} respectively. Sampling from the resulting posterior $P$ provides a policy with a bound on $C_\D(\pi)$ that holds with high probability (over the sampling of the training dataset $S$ and the policy $\pi$). 

Recent work has demonstrated the effectiveness of PAC-Bayes to provide strong bounds for deep neural networks \citep{Dziugiate17, Bartlett17, Perez-Ortiz20} and specifically for policy learning \citep{Majumdar18, Veer20, Ren20, Majumdar21}. However, the bounds used by these approaches do not provide a viable approach for performing OOD detection. The approaches are based on traditional PAC-Bayes bounds, where a \emph{distribution} $P$ over policies (e.g., a distribution over neural network weights) is chosen; the resulting bound is on $\EE_{\pi \sim P} C_\D(\pi)$ instead of $C_\D(\pi)$. Thus, given a test dataset $S'$ of environments, many policies from the distribution $P$ must be sampled in order to bound the expected cost on $S'$. This is not feasible in an OOD detection setting, where there is single execution on the test environments. Our use of the derandomized PAC-Bayes bound in Theorem~\ref{thm:derand_pacbayes_specialized} avoids this issue since we can bound $C_\D(\pi)$ for a \emph{particular} policy sampled from $P$.

We provide approaches for optimizing the bound provided in Theorem~\ref{thm:derand_pacbayes_specialized} using backpropagation (Appendix~\ref{ap:backprop alg}) and Evolutionary Strategies (ES) \citep{Wierstra14} (Appendix~\ref{app:reg-NES}). Since Theorem~\ref{thm:derand_pacbayes_specialized} requires a deterministic training algorithm, we fix the random seed for stochastic training methods. This makes the algorithm deterministic as the same input will always produce the same output. We choose multivariate Guassian distributions with diagonal covariance $\text{diag}(s)$, i.e., $P=\N(\mu,\text{diag}(s))$, for the posterior $P$ and prior $P_0$ distributions. Further, let $\psi:=(\mu,\log s)$; we use the shorthand $\N_\psi$ for $\N(\mu, \text{diag}(s))$. We denote $\pi_w$ with weights $w \sim \N_\psi$ as a parameterization of the robot's policy (e.g., neural networks with weights $w$). After training, we sample and fix a $w$ from the trained posterior for deployment on test environments. We then compute the PAC-Bayes upper bound $\overline{C}_\delta(\pi,S)$ and the PAC-Bayes lower bound $\underline{C}_\delta(\pi,S)$, each holding with probability $1-\delta$.

\subsection{Task-driven OOD detection with statistical guarantees}
\label{subsec:ood-detect-theory}

We now tackle the problem of OOD detection as defined in Section~\ref{sec:problem_formulation}. The PAC-Bayes training pipeline from Section~\ref{subsec:PAC-Bayes} produces a policy $\pi$ with associated bounds $\overline{C}_\delta(\pi,S)$ and $\underline{C}_\delta(\pi,S)$ on the expected cost $C_\D(\pi)$ that hold with probability $1-\delta$ over the sampling of the training dataset $S \sim \D^m$ and the policy $\pi \sim P$. Our key idea for OOD detection is that if our PAC bound $\overline{C}_\delta(\pi,S)$ is violated by $\pi$ in the test environments $S'$ (drawn from the test distribution $\D'$), then this indicates that the test environments are OOD-adverse and if $\underline{C}_\delta(\pi,S)$ is violated, then this indicates that the test environments are OOD-benign. We present two detectors below that formalize this intuition using two popular frequentist statistical inference tools --- hypothesis testing via p-values and confidence interval overlap.

\subsubsection*{Method 1: Hypothesis testing}

The first detector we present leverages hypothesis testing to declare one of the following three outcomes for the test dataset: OOD-adverse~($\oodRightSym$), OOD-benign~($\oodLeftSym$), or within distribution~(WD) by the detector. We perform this detection by computing upper bounds on the p-values that hold with high probability. Note that we do not make any normality assumption on the underlying distribution to estimate the p-values.

To perform hypothesis testing, we first establish a null-hypothesis $H_0$ and an alternate hypothesis $H_1$ which is the logical negation of $H_0$. Statistical inference is then performed by computing the p-value which is the likelihood of observing a test dataset $\hat{S}\sim \mathcal{D}'^n$ with an average cost more \emph{extreme}\footnote{We will check both, left and right, tails of the distributions.} than the average cost on the observed test dataset $S'\sim\mathcal{D}'^n$ assuming that $H_0$ holds. If the p-value drops below a significance level $\alpha\in(0,1)$, which is chosen before looking at the data, we can conclude that under the null-hypothesis the observed test dataset $S'$ had a very small probability of being drawn; therefore, the null-hypothesis $H_0$ can be rejected.

Our detector performs two hypothesis tests: (i) $H_0:\oodLeftSym$ and $H_1:\oodRightSym$ and (ii) $H_0:\oodRightSym$ and $H_1:\oodLeftSym$. If the first test returns a p-value smaller than the significance level $\alphaRight$, then we declare that the distribution $\mathcal{D}'$ from which the test dataset $S'$ is drawn is $\oodRightSym$ according to our task-driven notion \eqref{eq:task-driven OOD}; if the p-value of the second test is smaller than the significance level $\alphaLeft$, then we declare that environments drawn from the distribution $\mathcal{D}'$ are $\oodLeftSym$ according to our notion \eqref{eq:task-driven WD}. If both these tests are inconclusive, i.e., p-values for both are above the significance values, then we cannot declare either $\oodRightSym$ or $\oodLeftSym$ with confidence and therefore declare WD. A mathematically precise definition of the p-values for the two tests is given as follows.
\begin{definition}[adapted from \cite{Hogg05}]\label{def:p-value}
Let $\D'$ be the test distribution and $S'\sim\D'^n$ be an observed dataset. Let $\pi$ be the robot's control policy. Then, the p-value for $\oodRightSym$ detection is defined as: 
\begin{align}
    \pRight(S') &:= \underset{\hat{S}\sim\D'^n}{\mathbb{P}}[C_{\hat{S}}(\pi) \geq C_{S'}(\pi)~|~C_{\mathcal{D}'}(\pi) \leq C_\D(\pi)] ,
\end{align}
and the p-value for $\oodLeftSym$ detection is defined as:
\begin{align}
    \pLeft(S') &:= \underset{\hat{S}\sim\D'^n}{\mathbb{P}}[C_{\hat{S}}(\pi) \leq C_{S'}(\pi)~|~C_{\mathcal{D}'}(\pi) > C_\D(\pi)] .
\end{align}
\end{definition}

Since we lack an explicit form of the distributions $\mathcal{D}$ and $\mathcal{D}'$, direct computation of the p-values is not feasible. We alleviate this challenge by presenting upper bounds on the p-values by leveraging the PAC-Bayes generalization bounds (Theorem~\ref{thm:derand_pacbayes_specialized} and Corollary~\ref{cor:pac-bayes-lower-bound}). These upper bounds hold with probability $1-\delta$ (over the sampling of $S$ and $\pi$).

\begin{theorem}\label{thm:p-value}
Let $\D$ be the training distribution and $P$ be the posterior distribution on the space of policies obtained through the training procedure described in Section~\ref{subsec:PAC-Bayes}. Let $S'\sim\D'^n$ be a test dataset, $\pRight(S')$ and $\pLeft(S')$ be the p-values as defined in Definition~\ref{def:p-value}, $\deltaRight\in(0,1)$, and $\deltaLeft\in(0,1)$. Then,
\begin{align}\label{eq:p-value-bound}
    (i)~\underset{(S,\pi)\sim(\D^m\times P)}{\mathbb{P}} [\pRight(S') \leq \exp(-2n\overline{\tau}(S)^2)] \geq 1-\deltaRight , \\
    (ii)~\underset{(S,\pi)\sim(\D^m\times P)}{\mathbb{P}} [\pLeft(S') \leq \exp(-2n\underline{\tau}(S)^2)] \geq 1-\deltaLeft ,
\end{align}
where $\overline{\tau}(S):=\max\{C_{S'}(\pi) - \overline{C}_{\deltaRight}(\pi,S), 0\}$ and $\underline{\tau}(S):=\max\{ \underline{C}_{\deltaLeft }(\pi,S)-C_{S'}(\pi), 0\}$.
\begin{proof}
The proof is provided in Appendix~\ref{app:p-val-proof}.
\end{proof} 
\end{theorem}

Theorem~\ref{thm:p-value} provides an upper bound on the p-values which hold with high confidence. If the upper bound is below the respective significance levels $\alphaRight$ or $\alphaLeft$, then with high confidence we can say that the p-value is below $\alphaRight$ or $\alphaLeft$; thereby, Theorem~\ref{thm:p-value} facilitates OOD-adverse/OOD-benign detection through hypothesis testing. The resulting detector is detailed in the following algorithm:
\begin{algorithm}[h]
    \caption{OOD-adverse/OOD-benign Detection using Hypothesis Testing}
    \label{alg:HT-detector}
\begin{algorithmic}
    \State \textbf{Input}: $\deltaRight, \deltaLeft, \alphaRight, \alphaLeft  \in (0,1)$.
    \State \textbf{Input}: PAC-Bayes Bounds: $\overline{C}_{\deltaRight}(\pi,S)$, $\underline{C}_{\deltaLeft}(\pi,S)$.
    \State \textbf{Input}: Test dataset $S' \sim \D'^n$ and policy $\pi \sim P$.
    \State \textbf{Output}: $\oodRightSym$, $\oodLeftSym$, and WD
    \State $C_{S'}(\pi) \leftarrow \frac{1}{n} \sum_{E \in S'} C_E(\pi)$
    \State $\overline{\tau} \leftarrow \max\{C_{S'}(\pi) - \overline{C}_{\deltaRight}(\pi,S), 0\}$
    \State $\underline{\tau} \leftarrow \max\{\underline{C}_{\deltaLeft}(\pi,S) - C_{S'}(\pi) , 0\}$
    \If{$\exp(-2n\overline{\tau}^2) \leq \alphaRight$} 
        \State $\oodRightSym\leftarrow$ True
    \EndIf
    \If{$\exp(-2n\underline{\tau}^2) \leq \alphaLeft$} 
        \State $\oodLeftSym\leftarrow$ True
    \EndIf
    \If{$\exp(-2n\overline{\tau}^2) > \alphaLeft$ and $\exp(-2n\underline{\tau}^2) > \alphaRight$} 
        \State WD $\leftarrow$ True
    \EndIf
\end{algorithmic}
\end{algorithm}

A natural question to ask is whether the p-values for both hypothesis tests can be less than their respective significance levels $\alphaRight$ and $\alphaLeft$, implying that a dataset is simultaneously $\oodRightSym$ and $\oodLeftSym$ with high probability. In the forthcoming lemma we show that the above detector indeed returns \emph{mutually exclusive} outputs.
\begin{lemma}
\label{lem:p-val-consistent}
Algorithm~\ref{alg:HT-detector} returns mutually exclusive outputs, i.e., it returns only one of the three possibilities: $\oodRightSym$, $\oodLeftSym$, or WD.
\end{lemma}
\begin{proof} A detailed proof is provided in Appendix~\ref{app:p-val-consistent-proof}.
\end{proof}

\subsubsection*{Method 2: Confidence interval on the difference in expected train and test costs}
We now present another method for detecting task-relevant distribution shifts (Section~\ref{sec:problem_formulation}, Equation~\eqref{eq:task-driven OOD} and Equation~\eqref{eq:task-driven WD}) by providing bounds on the difference between expected test cost ($C_{\D'}(\pi)$) and the expected training cost ($C_{\D}(\pi)$) that hold with high probability. Using a confidence-interval-based method allows us to provide a guaranteed false positive and false negative rate for our detector, which is important for reliable use in safety-critical environments. We provide two lower bounds: (i) $\Delta \CRight$, which lower bounds $C_{\D'}(\pi) - C_{\D}(\pi)$ and (ii) $\Delta \CLeft$, which lower bounds $C_{\D}(\pi) - C_{\D'}(\pi)$. If $\Delta \CRight$ is positive then (with high confidence) $C_{\D'}(\pi) > C_{\D}(\pi)$, which corresponds to task-driven OOD-adverse detection. Similarly, if $\Delta \CLeft$ is non-negative then (with high confidence) $C_{\D}(\pi) \geq C_{\D'}(\pi)$, which corresponds to task-driven OOD-benign detection. Finally, if $\Delta \CLeft$ and $\Delta \CRight$ are negative then we cannot declare either OOD-benign ($\oodLeftSym$) or OOD-adverse ($\oodRightSym$) with confidence and therefore declare that environments drawn from the given test dataset $S'$ is within-distribution (WD). We formalize these high-confidence bounds in Theorem~\ref{thm:CI-all}.
\begin{theorem}\label{thm:CI-all}
Let $\D$ be the training distribution, $\D'$ be the test distribution, and $P$ be the posterior distribution on the space of policies obtained through the training procedure described in Section~\ref{subsec:PAC-Bayes}. Let $\deltaRight,\deltaRight'\in(0,1)$ such that $\deltaRight+\deltaRight' < 1$, $\gamma_A := \sqrt{\frac{\ln{(1/\deltaRight')}}{2n}}$, and $\Delta \CRight:= C_{S'}(\pi) - \gamma_A - \overline{C}_{\deltaRight}(\pi,S)$. Similarly, let $\deltaLeft,\deltaLeft'\in(0,1)$ such that $\deltaLeft+\deltaLeft' < 1$, 
$\gamma_B := \sqrt{\frac{\ln{(1/\deltaLeft')}}{2n}}$, and $\Delta \CLeft:= \underline{C}_{\deltaLeft}(\pi,S) - C_{S'}(\pi) - \gamma_B$. Then, 
\begin{align}
    (i)~\underset{(S,\pi,S')\sim(\D^m\times P \times \D'^n)}{\mathbb{P}}[C_{\D'}(\pi)& - C_{\D}(\pi) \geq \Delta \CRight ] \nonumber \\ & \geq 1-\deltaRight-\deltaRight' , \\
    (ii)~\underset{(S,\pi,S')\sim(\D^m\times P \times \D'^n)}{\mathbb{P}}[C_{\D}(\pi)& - C_{\D'}(\pi) \geq \Delta \CLeft ] \nonumber \\ & \geq 1-\deltaLeft-\deltaLeft' .
\end{align}
\begin{proof}
A detailed proof of this theorem is provided in Appendix~\ref{app:CI-proof}.
\end{proof}
\end{theorem}
The detection scheme based on Theorem~\ref{thm:CI-all} is outlined in Algorithm~\ref{alg:CI-detector}. It is important to note that with this detection scheme, the user can pick the desired false positive and false negative OOD detection rates by selecting $\deltaRight$, $\deltaRight'$, $\deltaLeft$, and $\deltaLeft'$ (see Remark~\ref{rem:fp_fn_guarantee}). This allows us to tune the detector's sensitivity to distribution shifts according to the situation in which it is deployed. For example, in safety-critical situations one may want to deploy a policy only when we are confident that the robot is operating OOD-benign or WD, hence, we can choose a low \emph{maximum permissible} false negative rate. However, when operating in non-safety-critical settings, a higher false negative rate can be tolerated, in which case the OOD-benign detector can afford to make declarations less cautiously.
\begin{algorithm}[ht]
    \caption{OOD-adverse/OOD-benign Detection using Confidence Intervals}
    \label{alg:CI-detector}
\begin{algorithmic}
    \State \textbf{Input}: $\deltaRight, \deltaRight' \in (0,1)$ with desired \textbf{maximum} false positive rate $\deltaRight + \deltaRight' < 1$. 
    \State \textbf{Input}: $\deltaLeft, \deltaLeft' \in (0,1)$ with desired \textbf{maximum} false negative rate, $\deltaLeft + \deltaLeft' < 1$. 
    \State \textbf{Input}: PAC-Bayes Bounds: $\overline{C}_{\deltaRight}(\pi,S)$, $\underline{C}_{\deltaLeft}(\pi,S)$.
    \State \textbf{Input}: Test dataset $S' \sim \D'^n$ and policy $\pi \sim P$.  
    \State \textbf{Output}: $\oodRightSym$, $\oodLeftSym$ and WD.
    \State $C_{S'}(\pi) \leftarrow \frac{1}{n} \sum_{E' \in S'} C_{E'}(\pi)$
    \State $\gamma_A \leftarrow \sqrt{\frac{\ln{(1/\deltaRight')}}{2n}}$
    \State $\gamma_B \leftarrow \sqrt{\frac{\ln{(1/\deltaLeft')}}{2n}}$
    \State $\Delta \CRight \leftarrow C_{S'}(\pi) - \gamma_A - \overline{C}_{\deltaRight}(\pi,S)$
    \State $\Delta \CLeft \leftarrow \underline{C}_{\deltaLeft}(\pi,S) - C_{S'}(\pi) - \gamma_B$
    \If{$\Delta \CRight > 0$} 
        \State $\oodRightSym$ $\leftarrow$ True
    \EndIf
    \If{$\Delta \CLeft \geq 0$} 
        \State $\oodLeftSym$ $\leftarrow$ True
    \EndIf
    \If{$\Delta \CRight \leq 0$ and $\Delta \CLeft < 0$} 
        \State WD $\leftarrow$ True
    \EndIf
\end{algorithmic}
\end{algorithm}

Similar to the detector presented in Algorithm~\ref{alg:HT-detector}, this detector also generates mutually exclusive outputs, i.e., it is impossible for it to detect both OOD-adverse and OOD-benign for the same test dataset; we formalize this below.
\begin{lemma}
\label{lem:CI-unambiguous}
Algorithm~\ref{alg:CI-detector} returns mutually exclusive outputs, i.e., it returns one of three possibilities: $\oodRightSym$, $\oodLeftSym$, or WD.
\begin{proof} A detailed proof is provided in Appendix~\ref{app:CI-consistent-proof}.
\end{proof}
\end{lemma}

With the confidence-interval-based detector as outlined in Algorithm~\ref{alg:CI-detector} and Theorem~\ref{thm:CI-all}, we can guarantee the false positive rate and the false negative rate of the detector to be upper bounded by $\deltaRight + \deltaRight'$ and $\deltaLeft + \deltaLeft'$ respectively, as shown in Remark~\ref{rem:fp_fn_guarantee}.

\begin{remark}
\label{rem:fp_fn_guarantee}
    The detection scheme presented in Algorithm~\ref{alg:CI-detector} has a false positive rate upper bounded by $\deltaRight + \deltaRight'$ and a false negative rate upper bounded by $\deltaLeft + \deltaLeft'$. This is evident from Theorem~\ref{thm:CI-all}, from which we know:
    \begin{enumerate}
        \item $\Delta \CRight > 0 \implies {\mathbb{P}}[C_{\D'}(\pi) > C_{\D}(\pi)] \geq 1-\deltaRight-\deltaRight' \iff  {\mathbb{P}}[C_{\D}(\pi) \geq C_{\D'}(\pi)] < \deltaRight +\deltaRight'$. Therefore, when the detector declares OOD-adverse ($\Delta \CRight > 0$), the environments may be OOD-benign with a probability \emph{at most} $\deltaRight + \deltaRight'$, i.e., the \textbf{maximum} false positive rate associated with this detector is upper bounded by $\deltaRight + \deltaRight'$.
        \item $\Delta \CLeft \geq 0 \implies {\mathbb{P}}[C_{\D}(\pi) \geq C_{\D'}(\pi)] \geq 1-\deltaLeft-\deltaLeft' \iff  {\mathbb{P}}[C_{\D'}(\pi) > C_{\D}(\pi)] < \deltaLeft +\deltaLeft'$. Therefore, when the detector declares OOD-benign ($\Delta \CLeft \geq 0$), the environments may be OOD-adverse with a probability \emph{at most} $\deltaLeft + \deltaLeft'$, i.e., the \textbf{maximum} false negative rate associated with this detector is upper bounded by $\deltaLeft + \deltaLeft'$.
    \end{enumerate}
\end{remark}

\section{Examples}
\label{sec:examples}

We demonstrate the ability of our approach to perform task-driven OOD detection with guaranteed confidence bounds on two examples in both simulation and on hardware: a manipulator grasping a new set of objects and a drone navigating a new set of environments. For the navigation task, we compare our methods with popular OOD detection baselines. Our code is available at: \href{https://github.com/irom-lab/Task_Relevant_OOD_Detection/tree/extensions}{\texttt{https://github.com/irom-lab/Task\_Relevant\_ OOD\_Detection/tree/extensions}}, and videos of the experiments can be found at \href{https://youtu.be/jKye3A09le0}{\texttt{https://youtu.be/jK ye3A09le0}}.

\subsection{Robotic grasping}
\label{subsec:grasping}

\begin{figure*}[h]
\ifarxiv \vspace{0mm} \else \vspace{1mm} \fi
\centering
\subfigure[\hspace{-2mm}]
{
\includegraphics[width=0.265\textwidth]{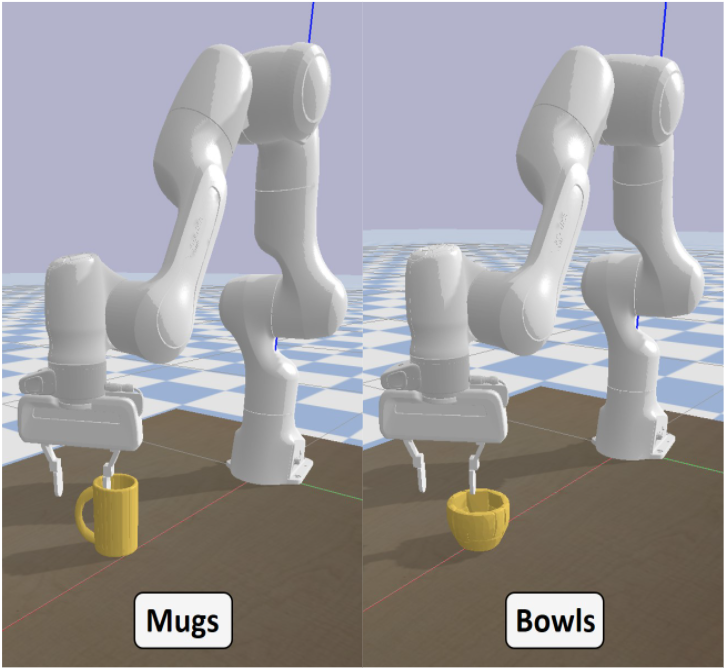}
\label{fig:franka-panda}
}
\hspace{-3mm}
\centering
\subfigure[\hspace{-5mm}]
{
\includegraphics[width=0.33\textwidth]{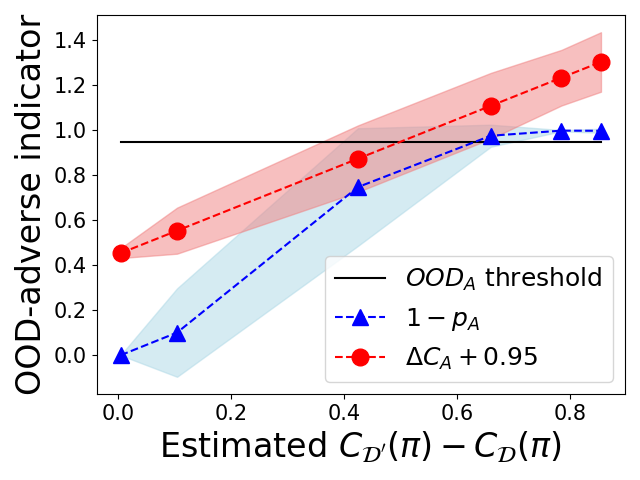}  
\label{fig:mug-pos}
}
\hspace{-4mm}
\subfigure[\hspace{-8mm}]
{
\includegraphics[width=0.33\textwidth]{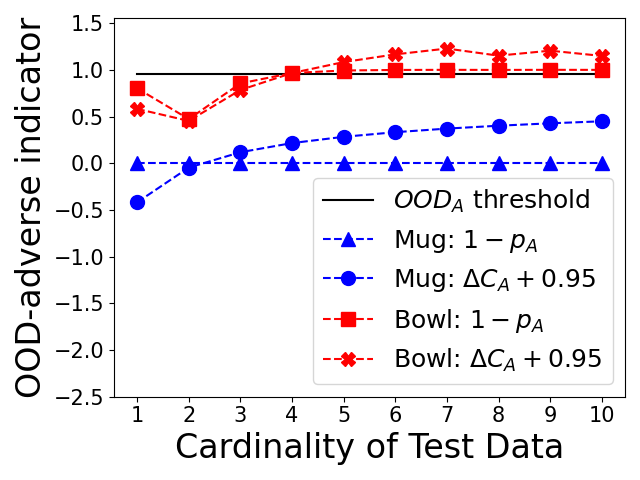}
\label{fig:mug-bowl}
}
\caption{OOD-adverse detection for grasping. \textbf{(a)} Franka Panda arm in PyBullet grasping a mug (left) and a bowl (right). \textbf{(b)} Performance of our OOD-adverse detectors for different distributions on mug placement, with the shaded region denoted the one standard deviation spread. Both our approaches perform similarly and the OOD-adverse indicators increase monotonically with $C_{\D'}(\pi)-C_\D(\pi)$. \textbf{(c)} Comparison of our OOD-adverse detectors for grasping mugs and grasping bowls. Both our approaches detect $\oodRightSym$ using a small number of test environments (just 4) for bowls and do not detect OOD-adverse for mugs (as expected). \label{fig:grasping-ood}}
\end{figure*}

\textbf{Overview.} We use the Franka Panda arm (Figure~\ref{fig:franka-panda}) for grasping objects in the PyBullet simulator \citep{Coumans18} and build upon the open-source code provided in \cite{Ren20}. The robot employs a vision-based control policy that uses a depth map of the object obtained from an overhead camera and returns an open-loop action $a:=(x,y,z,\theta)$ which corresponds to the desired grasp position and yaw orientation of the gripper. We train the manipulator to grasp mugs placed in $SE(2)$ poses drawn from a specific distribution. Then, we demonstrate the efficacy of our OOD-adverse detection framework by (i) gradually modifying the distribution on the mug poses and (ii) changing the objects from mugs to bowls. 

\textbf{Control policy.} The control policy is a deep neural network (DNN) which inputs a $128\times 128$ depth map of the object and a latent state $z\in\mathbb{R}^{10}$ sampled from a multivariate Gaussian distribution $\mathcal{N}_\psi$ with a diagonal covariance, and outputs an open-loop grasp action $a$; see Figure~\ref{fig:grasp-network} in Appendix~\ref{app:manipulator-exp} for the policy. In \cite{Ren20}, the distribution $\mathcal{N}_\psi$ on the latent space encodes prior domain/expert knowledge. 

\textbf{Training.} Mugs from the ShapeNet dataset \citep{Chang15} are randomly scaled in all dimensions to generate a training dataset $S$ of 500 mugs. If the robot is able to lift the mug by 10 cm, then we consider the rollout successful and assign a cost of 0; otherwise the cost is set to 1. In training, we optimize the distribution $\mathcal{N}_\psi$ on the latent space to minimize the PAC-Bayes upper bound provided in Theorem~\ref{thm:derand_pacbayes_specialized} using Algorithm~\ref{alg:pacbayes ES}, while the weights of the CNN and MLP networks in Figure~\ref{fig:grasp-network} in Appendix~\ref{app:manipulator-exp} remain fixed. The prior $\mathcal{N}_{\psi_0}$ is chosen as the normal distribution with zero mean and identity covariance. A policy $\pi$ is sampled from the trained posterior $\N_\psi$ and the PAC-Bayes bound for this policy is computed as $\overline{C}_\delta(\pi,S) = 0.1$ with $\delta = 0.01$. 

\subsection*{Simulation Results} 
We perform OOD-adverse and OOD-benign detection using the two methods presented in Theorem~\ref{thm:p-value} and Theorem~\ref{thm:CI-all}. For detection with p-value, we choose a significance level $\alpha_A = 95\%$, while, for detection using $\Delta C_{A}$ (the lower bound on $C_{\mathcal{D}'}-C_\mathcal{D}$) we choose a confidence level of $95\%$, i.e., $\delta_A+\delta'_A=0.05$, which ensures that the false-positive rate of our detector is no greater than $5\%$. We perform two experiments to demonstrate the efficacy of our approach. First, we make the distribution on the mug's initial placement progressively more challenging; see Appendix~\ref{app:manipulator-exp} for the exact distributions. For each distribution, we sample a test dataset of cardinality $10$  and compute our OOD indicators: (i) the lower bound on $1-p_A$ (where $p_A$ is the p-value) and (ii) $\Delta C_{A}$ using Theorem~\ref{thm:CI-all}. Figure~\ref{fig:mug-pos} plots the mean (dashed line) and a one standard deviation spread (shaded region) for the OOD indicators computed using 20 test datasets as a function of $C_{\mathcal{D}'} - C_\mathcal{D}$ (estimated via exhaustive sampling). Note that we plot $\Delta C_{A} + 0.95$ so that the OOD threshold is the same (0.95) for both methods. We compute the results for OOD-benign detection as well but do not report them here because we tested in settings that were more challenging than the training setting; thus, the OOD-adverse detector provided more interesting results. As the cost of the policy deteriorates on test distributions our $\oodRightSym$ indicators reliably increase, capturing the shift of the test distributions away from the training distribution. In the second experiment, we change the objects that the manipulator must grasp from mugs to bowls. Figure~\ref{fig:mug-bowl} shows that with a small test dataset $S'$ of cardinality 5, both our approaches detect $\oodRightSym$ when bowls are used (red curves). As expected, our OOD detectors are not triggered for mugs (blue curves), which are drawn from the training distribution.

\subsection*{Hardware results} 
\begin{figure*}[ht]
\centering
\subfigure[\hspace{-2mm}]
{
\includegraphics[width=0.25\textwidth]{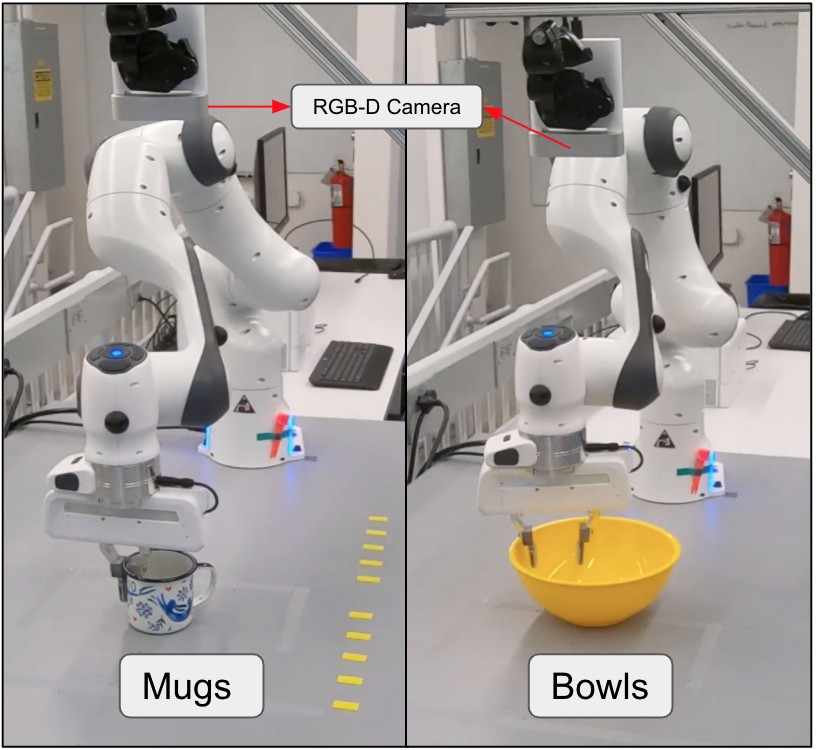}
\label{fig:franka-panda-hw}
}
\hspace{-3mm}
\subfigure[\hspace{-5mm}]
{
\includegraphics[width=0.30\textwidth]{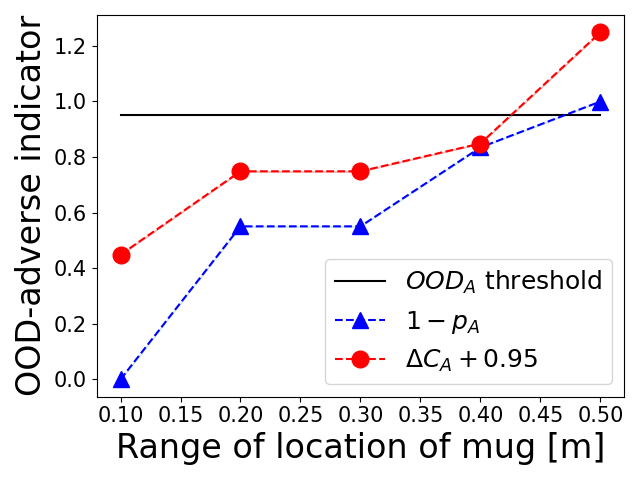}  
\label{fig:mug-pos-hw}
}
\vspace{-4mm}
\subfigure[\hspace{-8mm}]
{
\includegraphics[width=0.30\textwidth]{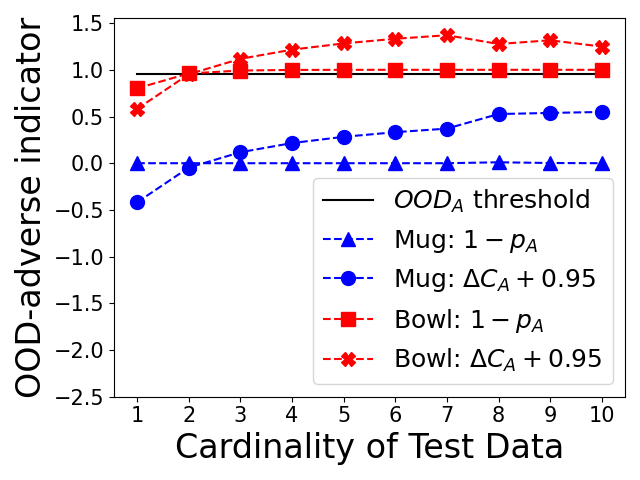}
\label{fig:mug-bowl-hw}
}
\caption{OOD detection for grasping. \textbf{(a)} Hardware Setup with camera mounted above the Franka Panda robot arm. \textbf{Left:} setup for varying mug locations, with (yellow) tape used to indicate the range of positions of the mug. \textbf{Right:} Attempted grasping of bowl \textbf{(b)} OOD detection for different distributions on mug placement, where the detector monotonically increases with increasing range in location. \textbf{(c)} Comparison of our OOD detector for grasping mugs and grasping bowls. We detect OOD using a small number of test environments (just 3) for bowls and do not detect OOD for mugs (as expected). \label{fig:grasping-ood-hw}}
\end{figure*}

We perform hardware experiments using the Franka Panda robot arm with input from a downward-facing camera mounted above the manipulator (Figure~\ref{fig:franka-panda-hw}). In these experiments, we use the same policy that was trained on mugs in simulation and then evaluate the performance of this policy on grasping 10 mugs with varied location (progressively encompassing a larger region) and grasping 10 bowls. Our experiments use mugs with shapes similar to those in the ShapeNet dataset \citep{Chang15}, to be consistent with the simulation. 

\textbf{Changing location of mugs.} For each mug we vary the location of the mugs according to the first 5 distributions outlined in Appendix~\ref{app:manipulator-exp}, which correspond to uniform distribution with ranges of 0.1m, 0.2m, 0.3m, 0.4m, and 0.5m centered around the middle of camera's field of view. The sixth distribution is neglected for the hardware trials, as it includes regions that were outside the field of view of the camera. As the range of location gets larger when compared with the training distribution, we find that the proportion of trials that the grasp fails (indicating a task relevant distribution shift) increases. This corresponds to the value of $\Delta C_{A} + 0.95$ and $1-p_A$ monotonically increasing as plotted in Figure~\ref{fig:mug-pos-hw}. This is consistent with the simulation results presented in Figure~\ref{fig:mug-pos}). In hardware, both our detectors declare $\oodRightSym$when the range of mug locations is 0.5 m, compared to the 0.4 m and 0.5 m ranges declared to be $\oodRightSym$ in simulation. 

\textbf{Grasping bowls.}
The value of $\Delta C_A + 0.95$ and $1-p_A$ after each attempted grasp of a bowl is plotted in Figure~\ref{fig:mug-bowl-hw}. In hardware, we detect $\oodRightSym$ with just 3 test environments of bowls which is similar to the 4 test environments needed in simulation. Overall, the hardware experiments are consistent with the $\oodRightSym$ detections in simulation where we detect $\oodRightSym$ when the range of mug locations increases and for a low-cardinality dataset of bowls.

\subsection{Vision-based obstacle avoidance with a drone}
\label{subsec:swing}

\textbf{Overview.} In both the simulation and hardware portions of this example, we aim to avoid an obstacle field with the Parrot Swing drone; this is an agile quadrotor/fixed-wing hybrid drone shown in Figure~\ref{fig:anchor}. We train a DNN control policy in a simulation setup based on the hardware system shown in Figure~\ref{fig:anchor}. The policy takes in a $50 \times 50$ depth image and outputs a softmax corresponding to a set of pre-computed motion primitives with the goal of avoiding obstacles by the largest distance. Since we designed the simulation portion of this example with application to hardware in mind, we have created motion primitives by capturing (with a Vicon motion tracking system) the trajectories of open-loop control inputs. This results in different maneuvers; the two images in Figure~\ref{fig:anchor} represent two of these trajectories. The use of motion primitives allows us to perform accurate sim-to-real transfer (as the motion primitives are recorded from the hardware system).

\begin{figure*}[h]
\centering
\subfigure[\hspace{-6mm}]{
\includegraphics[width=0.48\textwidth]{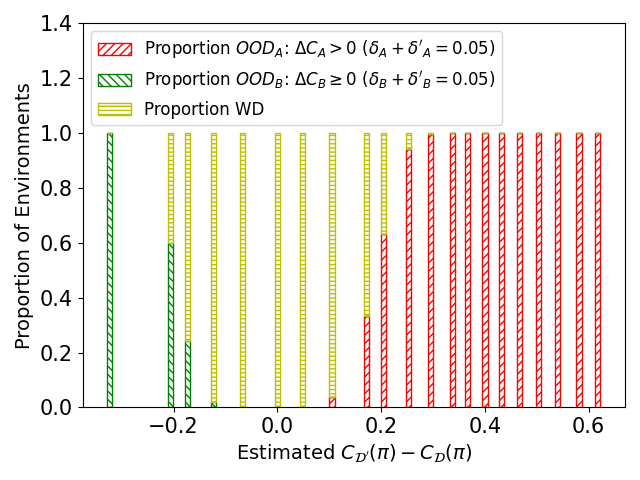}
\label{fig:complete_CI}}
\centering
\subfigure[\hspace{-9mm}]{
\includegraphics[width=0.48\textwidth]{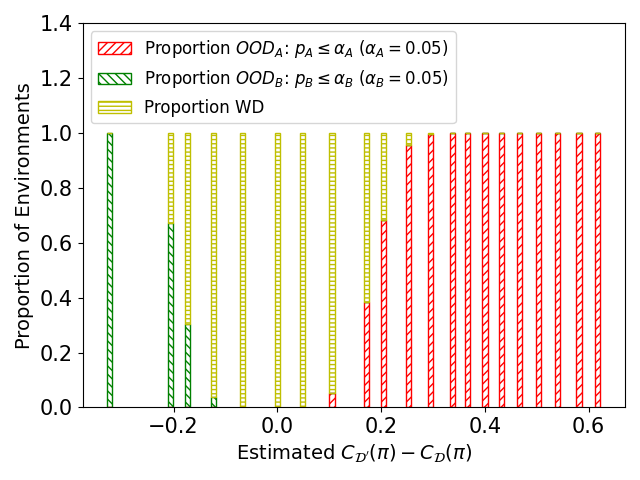}
\label{fig:complete_p}}
\vskip -6pt
\caption{The output of our detectors on settings with variable difficulty. \textbf{(a)} Confidence-interval based detector, \textbf{(b)} p-value based detector. Both detectors do not output any false declarations. When test and training costs vary significantly, our detectors declare OOD-adverse (red bars) and OOD-benign (green bars) with high confidence.} 
\label{fig:complete_detection}
\end{figure*}

\textbf{Training.} 
Environments consist of a set of randomly placed cylindrical obstacles. We record the minimum distance $d_\text{min}$ from the obstacles (as recorded by the robot's 120$^\circ$ field of view depth sensor) and assign a cost of $\max(0, 1 - \frac{d_\text{min}}{d_{thresh}})$ where $d_{thresh} = 100, 300, 500, 750, 1000$ mm. $d_{thresh}$ acts as a threshold radius around an obstacle beyond which the cost saturates to $0$. For each of these costs, we use $10{,}000$ training environments $S$ and train a prior to assign larger values to motion primitives which achieve a larger distance from obstacles. See Appendix~\ref{app:navigation-exp} for further details on the training procedure for the prior. We then use another $10{,}000$ environments to train the posterior distribution using Algorithm~\ref{alg:pacbayes backprop}. We sample a policy $\pi$ from the trained posterior and compute the PAC-Bayes bounds $\overline{C}_\delta(\pi, S)$ and $\underline{C}_\delta(\pi, S)$ with $\delta = 0.01$ for each of the datasets. 
\subsubsection*{Simulation Results}
To evaluate our OOD detection methods on vision-based obstacle avoidance in simulation, we randomly generate test datasets of varied environment difficulty by changing the number of obstacles and the maximum or minimum gap-size between obstacles. We perform OOD detection using 10 test environments for each difficulty setting (and present results averaged over 2000 such datasets). With these datasets, we estimate the expected test cost that the policy $\pi$ would incur on any given difficulty setting; this estimate is used to evaluate which environments our detection schemes \emph{should} declare as OOD-adverse/OOD-benign (i.e., which difficulty settings are $\oodRightSym$/$\oodLeftSym$ in our task-relevant sense). To estimate the test cost, we use $d_{thresh} = 500$ mm for the simulation results presented in this section. 
To evaluate our detectors in simulation, we first verify that the guarantees presented in Section~\ref{subsec:ood-detect-theory} do indeed hold. We then compare our detectors with two OOD detection baselines: (i) maximum softmax probability (MSP) \citep{Hendrycks17} (an effective and popular baseline for OOD detection), and (ii) MaxLogit \citep{Hendrycks20} (a recent state-of-the-art OOD detection baseline). We find that our detection schemes perform similarly to these baselines for task-relevant shifts. However, the baseline detectors are triggered by task irrelevant shifts, while our methods are only triggered by task-relevant shifts. Additionally, our detectors provide guarantees on the false-positive and false-negative rate and can declare WD environments and also differentiate between OOD-adverse and OOD-benign environments, while the baselines are limited to detecting OOD generically and provide no guarantees.

\textbf{OOD-adverse and OOD-benign detection.} To demonstrate our detection schemes (Algorithms~\ref{alg:HT-detector} and~\ref{alg:CI-detector}), we evaluate the proportion of datasets $S'\sim \D'^m$ from a given distribution that our detectors declare OOD-adverse, OOD-benign, or WD (Figure~\ref{fig:complete_detection}). The x-axis of this plot is the estimated cost (with $d_{thresh} = 500mm$) of the environment as compared to the training environment; positive $\Delta C_A$ indicate (task-relevant) $\oodRightSym$ environments (Equation~\ref{eq:task-driven OOD}) and non-negative values indicate (task-relevant) $\oodLeftSym$ environments (Equation~\ref{eq:task-driven WD}). Note that at 0 on the x-axis of this, we draw datasets from the training distribution. We observe that our confidence-interval based detector (Algorithm~\ref{alg:CI-detector}, Figure~\ref{fig:complete_CI}) maintains both its false positive and false negative rate guarantee, and does not declare \emph{any} $/oodRightSym$ environments as $/oodLeftSym$ or vice versa (see Figure~\ref{fig:swingsim_lowerbound} in Appendix~\ref{app:navigation-exp} for numerical validation of the guaranteed false positive rate). Additionally, the p-value based method (Algorithm~\ref{alg:HT-detector}, Figure~\ref{fig:complete_p}) performs very similarly and also does not incorrectly classify OOD-adverse as OOD-benign or vice versa. The accuracy of these declarations, however, comes at the cost of conservatism where both detectors output WD when $C_S'(\pi) - C_S(\pi)$ is small, i.e., environments which are on the border of OOD-adverse/OOD-benign. One can counteract this by changing $\delta'_A$ and $\delta'_B$ and therefore the desired guarantees. 

\begin{figure*}[h] 
\centering
\subfigure[\hspace{-6mm}]{
\includegraphics[width=0.48\textwidth]{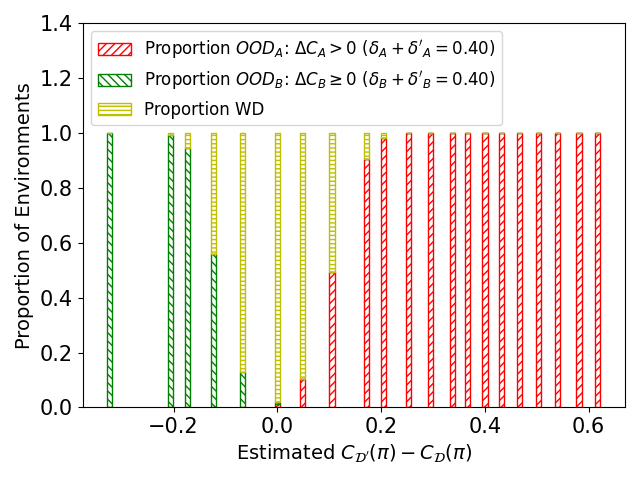}
\label{fig:changing_delta_less_conservative}}
\centering
\subfigure[\hspace{-9mm}]{
\includegraphics[width=0.48\textwidth]{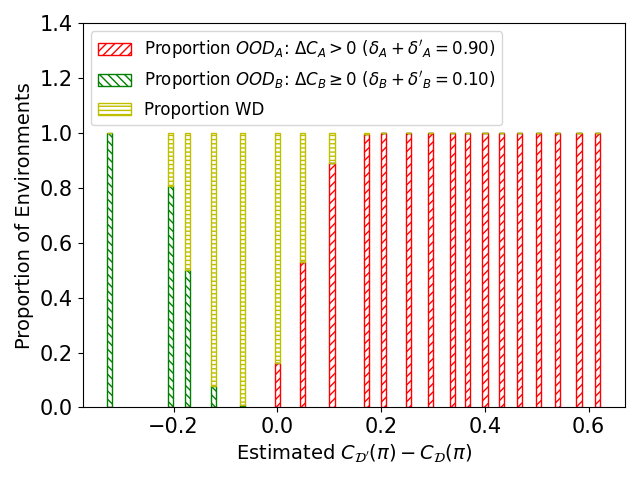}
\label{fig:changing_delta_safety_critical}}
\vskip -6pt
\caption{Investigating how changing the guaranteed false positive rate ($\delta'_A + \delta_A$) and false negative rate ($\delta'_B + \delta_B$) impacts the behavior of our confidence-interval based detector. \textbf{(a)} By increasing $\delta_A + \delta'_A$ and $\delta_B + \delta'_B$ to 40\% we reduce the number of within distribution declarations by making the detectors more sensitive to increases in cost (OOD-adverse detection) and decreases in the cost (OOD-benign detection). \textbf{(b)} A detector tuned for safety-critical contexts, where a low maximum false negative rate (10\%) is desired but a high maximum false-positive rate (90\%) may be permissible. This skews the detector to declare OOD-adverse for environments (red bars) with similar test and train costs.} 
\label{fig:changing_delta}
\end{figure*}

\textbf{Choosing desired maximum false-positive and false-negative rate.} The confidence-interval based detector (Algorithm~\ref{alg:CI-detector}), as shown in Remark~\ref{rem:fp_fn_guarantee}, allows for the desired \emph{maximum permissible} false positive rate ($\delta_A + \delta'_A$) and false negative rate ($\delta_B + \delta'_B$) to be picked a priori. Higher admissible false positive/negative rates would allow the detector to declare OOD-adverse/OOD-benign instead of WD for test environments with similar costs to training environments, i.e., for values of $C_D'(\pi) - C_D(\pi)$ close to zero. On the contrary, lower admissible false positive/negative rates would result in OOD-adverse/OOD-benign declarations only for those environments with test costs very different from training costs. This property could be especially useful in practice, where the threshold difference in test and train costs for which OOD-adverse/OOD-benign is declared could be altered depending on how safety critical the situation is. For exposition, we pick $\delta_A = \delta_B = 0.01$ and use only $\delta'_A$ and $\delta'_B$ to vary the maximum permissible false positive/negative rate. In Figure~\ref{fig:changing_delta_less_conservative}, we show that increasing our permissible false positive and false negative rates to 40\% ($ \delta_A + \delta'_A = \delta_B + \delta'_B  = 0.40$) results in WD declarations being replaced by OOD-benign or OOD-adverse declarations as compared to the 5\% false positive/negative rate in Figure~\ref{fig:complete_CI}. It is also possible to set $\delta'_A$ and $\delta'_B$ to be different values, as is displayed in Figure~\ref{fig:changing_delta_safety_critical}, where the permissible false negative rate is chosen to be 10\% ($\delta_B + \delta'_B = 0.10$), while the permissible false positive rate is 90\% ($\delta_A + \delta'_A = 0.90$). This is particularly suitable for safety-critical contexts where it can be dangerous to declare OOD-adverse environments as OOD-benign, i.e., a low false negative rate is desirable. In Figure~\ref{fig:changing_delta_safety_critical}, we see that we declare OOD-adverse (red bars) for more environments, including some of those with similar test and train costs and only declare OOD-benign when the test cost is significantly below the training cost. Note that the true false positive/negative rate is below the maximum permissible rates, consistent with the guarantee presented in Remark~\ref{rem:fp_fn_guarantee}. 

\begin{figure}[ht]
\centering
\includegraphics[width=1.0\columnwidth]{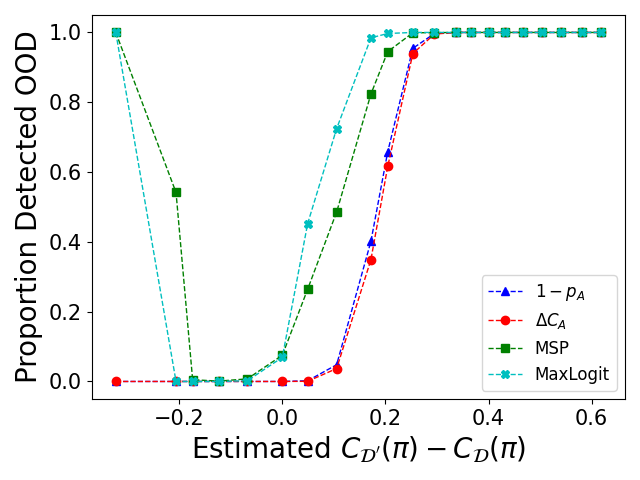}
\caption{Comparison of the performance of our OOD detectors with baselines MSP \citep{Hendrycks17} and MaxLogit \citep{Hendrycks20} on settings of variable difficulty (with $d_{thresh} = 500$mm). The $1-p_{A}$ and $\Delta C_{A}$ methods do not detect OOD when $C_{\D'}(\pi) < C_\D(\pi)$ whereas the baselines detect these task-irrelevant shifts in the environment. For $C_{\D'}(\pi) > C_\D(\pi)$ the baselines detect a higher proportion of datasets as OOD for smaller values of $C_{\D'}(\pi)-C_\D(\pi)$.}
\label{fig:swingsim_compare}
\end{figure}

\textbf{Comparison with Baselines for OOD detection.} We compare our task-driven OOD-adverse detection approach with two baselines: (i) maximum softmax probability (MSP) \citep{Hendrycks17}, and (ii) MaxLogit \citep{Hendrycks20}. We note that these baselines are specifically designed for networks which output categorical distributions, and thus provide strong benchmarks. It is important to note that these baselines do not provide a means to detect OOD-benign environments and so to compare our detection methods to these baselines, we only consider our OOD-adverse declarations. The results are plotted in Figure~\ref{fig:swingsim_compare}, using a p-value of $0.05$ and a guaranteed false-positive and false-negative rate of $5\%$ for the confidence interval method. We see that these guarantees do indeed hold, and declare a OOD-benign environments as OOD-adverse less than 5\% of the time. In contrast, the baselines do not provide any guarantees; they violate the false positive rate even on new environments drawn from the training distribution ($C_D'(\pi) - C_D(\pi) = 0$). Additionally, the baselines do not differentiate between OOD-adverse and OOD-benign environments; they declare environments with $C_D(\pi) > C_D'(\pi)$ as OOD, when in practice the policy would not result in failure on these (easier) environments. With regard to OOD-adverse environments, our detectors (which perform similarly) only detect test environments with higher expected costs as OOD-adverse, demonstrating that it is task-relevant. 

\textbf{Task-irrelevant shift.} As seen in Figure~\ref{fig:swingsim_compare}, the baselines are triggered even for task-irrelevant shifts in the environment distribution. For example, the baselines are triggered on environments with lower expected costs than the training environment ($C_{D'}(\pi) - C_D(\pi) = -0.3$). To investigate this further, we compare the baselines with our methods on a distribution where environments consist of $4$ (uniformly) randomly located obstacles, and we evaluate costs using a threshold distance, $d_{thresh} = 500$mm. In this setting, the control policy achieves a near-identical expected cost $C_{\mathcal{D}'}(\pi)$ (as estimated by exhaustive sampling of environments) to the expected training cost $C_{\mathcal{D}}(\pi)$ (in particular, $C_{\mathcal{D}'}(\pi)$ - $C_{\mathcal{D}}(\pi) = -0.02$). For this setting, MSP \citep{Hendrycks17} classified 100\% of test datasets as OOD and MaxLogit \citep{Hendrycks20} classified 98.6\% as OOD. Thus, the baselines are triggered by a task-irrelevant shift in the distribution. In contrast, our OOD-adverse detection method had a detection rate of only $0.1\%$ in this setting.

\subsubsection*{Hardware Results}
We use a Parrot Swing drone for the hardware experiments (Figure~\ref{fig:anchor}). We simulate a depth sensor for the drone (as if the sensor was mounted on the drone) by generating a synthetic depth image using the positions of objects from the Vicon motion capture system. We do not provide any other information to the policy, such as the position of obstacles or the environmental wind conditions. We generate each environment the same way as in simulation, and then place the real-world obstacles in the generated locations.

\begin{figure}[ht]
\centering
\includegraphics[width=1.0\columnwidth]{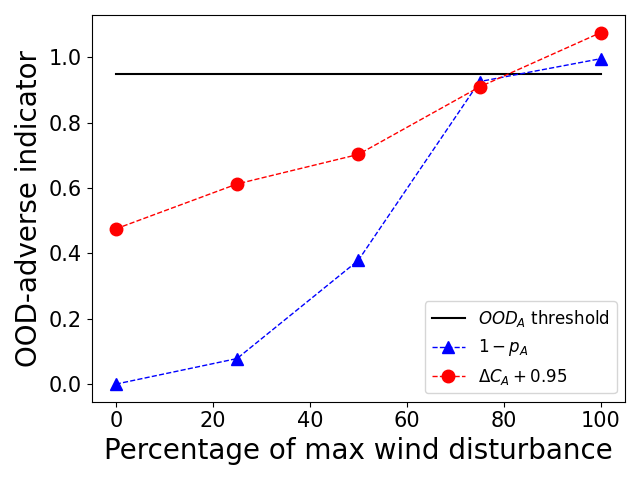}
\caption{Comparison of our OOD-adverse detectors on the Parrot Swing hardware for increasing wind disturbance. Both OOD indicators increase monotonically with $C_{\D'}(\pi)-C_\D(\pi)$ and are able to detect OOD-adverse at $100\%$ wind. Note that the costs are computed using $d_{thresh} = 300$ mm for the hardware results.}   
\label{fig:swinghardware}
\end{figure}

\textbf{Varied environment difficulty and wind disturbances.}
We deploy the policy trained in simulation on three kinds of OOD environments in hardware: (i) environments with a smaller number of obstacles (i.e., ``easier" environments), (ii) environments with smaller gaps between obstacles (i.e., ``harder" environments), and (iii) environments with wind generated using a fan (Figure~\ref{fig:anchor} right) with the same obstacle distribution as training. For each setting, we run $10$ trials on the hardware and use this for OOD-adverse detection, where we evaluate costs using the obstacle threshold radius, $d_{thresh} = 300$ mm. As expected, our OOD-adverse detectors are not triggered by the easier environments. For the harder environments, we compute $1-p \geq 0.81$ and $\Delta C_{\mathrm{fp}} = -0.11$. Results from the windy environments are shown in Figure~\ref{fig:swinghardware} for increasing values of wind (up to about $5~\mathrm{m/s}$). We note that the sim-to-real distribution shift (corresponding to the zero wind case) is not viewed as being OOD in a task-relevant manner by our approaches. Both our approaches assign OOD-adverse with increasingly high confidence as the wind speed is increased. We note that this detection is despite the fact that disturbances such as wind cannot be detected via the depth image given to the robot's policy. Thus any OOD detection technique which relies solely on the output of the policy, such as MSP \citep{Hendrycks17} and MaxLogit \citep{Hendrycks20}, would be unable to detect these environments as OOD. 

\section{Conclusion}
\label{sec:conclusion}

We have presented a framework for performing task-driven OOD detection with statistical guarantees. Our approach uses PAC-Bayes theory to train a policy with a bound on the expected cost on the training distribution. We then perform OOD-adverse and OOD-benign detection on test environments by checking for violations of the bound using approaches based on both p-values and confidence intervals. Both approaches provide strong performance for OOD detection; however, the approach based on p-values performs marginally better. Despite slightly worse empirical performance, the confidence interval approach is able to provide guarantees on the maximum false-negative and false-positive rate, unlike the approach based on p-values. Our simulated and hardware experiments demonstrate the ability of our approaches to perform OOD detection within a handful of trials. Comparisons with baselines also demonstrate two advantages: our OOD detectors (i) are sensitive to task-relevant distribution shifts, and (ii) provide statistical guarantees on detection. Additionally, we demonstrate the ability to tune our detectors' sensitivity by varying the maximum permissible false positive/negative rate. In particular, we show that increasing the maximum false positive rate makes our detector more sensitive to OOD-adverse environments and therefore, particularly well suited for safety-critical environments. Finally, we validate these results by deploying our detectors on hardware and demonstrate the effectiveness of our detectors on a vision-based navigation task as well as a manipulation task.

{\bf Challenges and future work.} It would be of practical interest to extend our approach to settings where the robot encounters environments in an \emph{online} manner (instead of the batch setting we consider here). Another particularly exciting direction is to develop versions of our approach that are more \emph{proactive}; instead of having to incur costs on the test environments, one could potentially perform OOD-adverse detection based on \emph{predicted} costs (thus avoiding the need to potentially fail on the test environments). Finally, another potential direction for future research is to leverage the OOD-adverse/OOD-benign detection schemes presented here to learn policies that are \emph{distributionally robust}; in particular, one could envision learning policies such that environments from a broad range of test distributions are detected as OOD-benign in the task-relevant sense employed in this paper. 

\subsubsection*{Acknowledgments}
The authors were supported by the Office of Naval Research [N00014-21-1-2803, N00014-18-1-2873], the NSF CAREER award [2044149], and the Toyota Research Institute (TRI). This article solely reflects the opinions and conclusions of its authors and not ONR, NSF, TRI or any other Toyota entity.

\bibliographystyle{IEEEtran}
\bibliography{bib.bib}

\newpage
\appendix
\section{Appendix}
\subsection{Proof of Theorem~\ref{thm:derand_pacbayes_specialized}}
\label{ap:proof derand pacbayes}

We begin with an introduction to the PAC-Bayes framework, and then provide a complete proof of Theorem~\ref{thm:derand_pacbayes_specialized}. PAC-Bayes provides an upper bound on the expected cost of deploying a policy distribution $P$ on environments $E$ drawn from an unknown distribution $\mathcal{D}$, i.e., $\EE_{E \sim \D} \EE_{\pi \sim P} C_E(\pi)$. This upper bound only depends on the cost of deploying $P$ in a finite set of training environments $S\sim\D^m$, i.e., the training cost $\EE_{\pi \sim P} C_{E_i}(\pi)$, and a regularizer which depends on the KL-divergence between $P$ and a prior $P_0$ that is chosen before observing $S$; note that $P_0$ need not be a Bayesian prior. The following is the PAC-Bayes bound that was presented in \cite{McAllester99} and tightened in \cite{Maurer04}:
\begin{theorem}[PAC-Bayes Bound \citep{McAllester99}] For any distribution over environments $\D$, data-independent prior distribution $P_0$, cost $C$ bounded in $[0,1]$, $m \geq 8$, and $\delta \in (0,1)$, with probability at least $1-\delta$ over a sampling of $S \sim \D^m$, the following holds for all posterior distributions $P$: \label{thm:mcallester}
\begin{align}
    \underset{E \sim \D}{\EE} \ \underset{\pi \sim P}{\EE} \ C_E(\pi) \leq & \sum_{i=1}^n \underset{\pi \sim P}{\EE} \ C_{E_i}(\pi) \\ & + \sqrt{\frac{\KL(P \| P_0) + \ln\frac{2\sqrt{m}}{\delta}}{2m}} \nonumber
\end{align} 
where $\KL$ is the KL-divergence. \end{theorem}
The above theorem and the forthcoming PAC-Bayes theorems in this section are presented for policy learning instead of supervised learning using the reduction provided in \cite{Majumdar18}. Note that this bound provides a guarantee for a \emph{distribution} over policies rather than a specific policy. This allows for a regularizer dependent on the KL-divergence between the prior and posterior distributions rather than one which is a direct expression of the complexity of the policy space (such as the VC-dimension). However, this creates a challenge for calculating the upper bound, which requires computing an expectation over $\pi \sim P$, or using potentially-loosening sample convergence bounds. Thus, we make use of the recent work which provides a framework for derandomized PAC-Bayes bounds (i.e. bounds which hold for a sampling of policy $\pi$ rather than an expectation over $\pi \sim P$) \citep{Viallard21}. The following is a general theorem for formulating the derandomized PAC-Bayes bounds:

\begin{theorem}[Pointwise PAC-Bayes Bound \citep{Viallard21}]
\label{pointwise_pacbayes}
For any positive function $\phi$, distribution $\D$, prior distribution $P_0$, and $\delta \in (0,1)$, with probability $1-\delta$ over a sampling of $S \sim \D^m$ and $\pi \sim P$, the following holds for any posterior distribution $P$:
\begin{align}
    \frac{\alpha}{\alpha - 1}&\ln(\phi(\pi, S)) \leq D_\alpha(P\| P_0) \ + \\ & \ln\bigg{(}\frac{1}{(\delta/2)^{\frac{\alpha}{\alpha - 1} + 1}} \underset{S'\sim\D^m}{\EE} \underset{\pi' \sim P_0}{\EE}\phi(\pi', S')^{\frac{\alpha}{\alpha - 1}} \bigg{)}  \nonumber
\end{align}
where $P$ is the output of algorithm $A$ on the training data $S$, i.e. $P:=A(P_0, S)$ and $D_\alpha$ is the R\'enyi divergence. 
\end{theorem}
Now we can proceed with the statement and proof.

\begin{customthm}{1}
For any distribution $\D$, prior distribution $P_0$, $\delta \in (0,1)$, cost bounded in $[0,1]$, and deterministic algorithm $A$ which outputs the posterior distribution $P$  we have the following:
\begin{align}
    \underset{(S, \pi) \sim (\D^m \times P)}{\PP}\ &\Bigg[ C_\D(\pi) \leq C_S(\pi) + \sqrt{\frac{D_2(P \| P_0) + \ln\frac{2\sqrt{m}}{(\delta/2)^3}}{2m}}\Bigg]\nonumber \\ & \leq 1 - \delta
\end{align} 
where $D_2$ is the R\'enyi Divergence for $\alpha = 2$.
\begin{proof} We begin with the statement in Theorem~\ref{pointwise_pacbayes}, which is proved in \cite{Viallard21}. Let $\alpha = 2$ and $\phi(\pi, S) = \exp[\frac{\alpha-1}{\alpha}m \KL(C_S(\pi) \| C_\D(\pi))]$. Thus, we have the following with at least probability $1 - \delta$ over the random choice $S \sim \D^m$ and $\pi \sim P$:
\begin{align}
    \KL&(C_S(\pi)\| C_\D(\pi)) \leq \frac{1}{m}\bigg{[}D_2(P \| P_0) \ + \\ 
    & \ln\bigg{(}\frac{1}{(\delta/2)^3} \ \underset{S'\sim\D^m}{\EE} \ \underset{\pi' \sim P_0}{\EE} \ e^{m\KL(C_{S'}(\pi') \| C_\D(\pi'))}\bigg{)} \bigg{]}  \nonumber
\end{align}
From \cite{Maurer04}, we can upper bound $\EE_{S'\sim\D^m}\  \EE_{\pi' \sim P_0} \ e^{m\KL(C_{S'}(\pi') \| C_\D(\pi'))}$ by $2\sqrt{m}$ when $m \geq 8$. This gives us the following bound
\begin{equation}
    \KL(C_S(\pi)\| C_\D(\pi)) \leq \frac{1}{m}\bigg{[}D_2(P \| P_0) + \ln\frac{2\sqrt{m}}{(\delta/2)^3}\bigg{]}.
\end{equation} 
We then apply the Pinkser's inequality, i.e. $\KL(p\|q) \leq c \implies q \leq p + \sqrt{c/2}$, which results in Inequality \eqref{pointwise_pacbayes_specialized}. Note that we could also use a quadratic version of the upper bound for the KL divergence between two distributions and produce an upper bound analogous to the one presented in \cite{Rivasplata19}. 
\end{proof}
\end{customthm}

\subsection{Proof of Theorem~\ref{thm:p-value}}
\label{app:p-val-proof}

For the readers' convenience, we restate Theorem~\ref{thm:p-value} here and provide a detailed proof.

\begin{customthm}{\ref{thm:p-value}}
Let $\D$ be the training distribution and $P$ be the posterior distribution on the space of policies obtained through the training procedure described in Section~\ref{subsec:PAC-Bayes}. Let $S'\sim\D'^n$ be a test dataset, $p_A(S')$ and $p_B(S')$ be the p-values as defined in Definition~\ref{def:p-value}, $\delta_A\in(0,1)$, and $\delta_B\in(0,1)$. Then,
\begin{align}
    & (i)~\underset{(S,\pi)\sim(\D^m\times P)}{\mathbb{P}} [p_A(S') \leq \exp(-2n\overline{\tau}(S)^2)] \geq 1-\delta_A , \\
    &(ii)~\underset{(S,\pi)\sim(\D^m\times P)}{\mathbb{P}} [p_B(S') \leq \exp(-2n\underline{\tau}(S)^2)] \geq 1-\delta_B  .
\end{align}
where $\overline{\tau}(S):=\max\{C_{S'}(\pi) - \overline{C}_{\delta_A}(\pi,S), 0\}$ and $\underline{\tau}(S):=\max\{ \underline{C}_{\delta_B} (\pi,S)-C_{S'}(\pi), 0\}$.
\end{customthm}

To prove Theorem~\ref{thm:p-value}, we establish the following lemmas.

\begin{lemma}\label{lem:p-val-O}
Let the assumptions of Theorem~\ref{thm:p-value} hold. For notational simplicity and without loss of generality, we let $\delta_{A} = \delta$. Then,
\begin{align}
    \underset{(S,\pi)\sim(\D^m\times P)}{\mathbb{P}} [p_A(S') \leq \exp(-2n\overline{\tau}(S)^2)] \geq 1-\delta ,
\end{align}
where $\overline{\tau}(S):=\max\{C_{S'}(\pi) - \overline{C}_\delta(\pi,S), 0\}$.
\end{lemma}
\begin{proof}
We prove this lemma by considering two cases: when the PAC-Bayes cost inequality in Theorem~\ref{thm:derand_pacbayes_specialized} holds, i.e., $C_\D(\pi) \leq \overline{C}_\delta(\pi,S)$, and when it does not, i.e., $C_\D(\pi) > \overline{C}_\delta(\pi,S)$; the two cases are considered in \eqref{eq:p-val-cond-begin}-\eqref{eq:p-val-cond}. In the latter case, we cannot say anything about the p-value, while in the former case, which holds with probability at least $1-\delta$, we show in \eqref{eq:claim}-\eqref{eq:thm-1-end} that $p_A(S') \leq \exp(-2n\overline{\tau}(S)^2)$.

Let us begin the proof by conditioning $\underset{(S,\pi)\sim(\D^m\times P)}{\mathbb{P}} [p_A(S') \leq \exp(-2n\overline{\tau}(S)^2)]$ as follows:
\small
\begin{align}
    \underset{(S,\pi)\sim(\D^m\times P)}{\mathbb{P}} [&p_A(S') \leq \exp(-2n\overline{\tau}(S)^2)] \label{eq:p-val-cond-begin} = \nonumber \\ 
    \underset{(S,\pi)\sim(\D^m\times P)}{\mathbb{P}} [&p_A(S') \leq \exp(-2n\overline{\tau}(S)^2)~|~C_{\D}(\pi)\leq \overline{C}_\delta(\pi,S)] \nonumber \\ & \underbrace{\underset{(S,\pi)\sim(\D^m\times P)}{\mathbb{P}}[C_\D(\pi) \leq \overline{C}_\delta(\pi,S)]}_{\geq 1-\delta~\text{(from~Theorem~\ref{thm:derand_pacbayes_specialized})} } +  \nonumber \\
    & \hspace{-2cm} \underbrace{\underset{(S,\pi)\sim(\D^m\times P)}{\mathbb{P}} [p_A(S') \leq \exp(-2n\overline{\tau}(S)^2)~|~C_{\D}(\pi)> \overline{C}_\delta(\pi,S)]}_{\geq 0} \nonumber \\ & \underbrace{\underset{(S,\pi)\sim(\D^m\times P)}{\mathbb{P}}[C_\D(\pi) > \overline{C}_\delta(\pi,S)]}_{\geq 0} \geq \\
    \underset{(S,\pi)\sim(\D^m\times P)}{\mathbb{P}} [&p_A(S') \leq \exp(-2n\overline{\tau}(S)^2)~|\nonumber \\ & C_{\D}(\pi)\leq \overline{C}_\delta(\pi,S)] (1-\delta) \label{eq:p-val-cond}
\end{align}
\normalsize
Now, we claim:
\begin{align}
    \underset{(S,\pi)\sim(\D^m\times P)}{\mathbb{P}} [& p_A(S') \leq \exp(-2n\overline{\tau}(S)^2)~|\nonumber \\ & C_{\D}(\pi)\leq \overline{C}_\delta(\pi,S)] = 1, \label{eq:claim}
\end{align}
which on using in \eqref{eq:p-val-cond} completes the proof of this lemma. The remainder of this proof is dedicated to establishing the claim in \eqref{eq:claim}.

We are given 
\begin{align}\label{eq:assump-1}
    C_{\D}(\pi)\leq \overline{C}_\delta(\pi,S) .
\end{align}
From Definition~\ref{def:p-value}, we have
\begin{align}
    p_A(S') =& \ \underset{\hat{S}\sim\D'^n}{\mathbb{P}}[C_{\hat{S}}(\pi) \geq C_{S'}(\pi)~|~C_{\D'}(\pi) \leq C_\D(\pi)] \\
    = \underset{\hat{S}\sim\D'^n}{\mathbb{P}}[&C_{\hat{S}}(\pi)-\overline{C}_\delta(\pi,S) \geq C_{S'}(\pi) - \overline{C}_\delta(\pi,S)~|\nonumber \\ & C_{\D'}(\pi) \leq C_\D(\pi)] \\ 
    = \underset{\hat{S}\sim\D'^n}{\mathbb{P}}[&C_{\hat{S}}(\pi)-\overline{C}_\delta(\pi,S) \geq \overline{\tau}~|~C_{\D'}(\pi) \leq C_\D(\pi)] \label{eq:bound-0} .
\end{align}
From \eqref{eq:assump-1} and the assumption that the null hypothesis holds in \eqref{eq:bound-0}, it follows that $C_{\D'}(\pi) \leq \overline{C}_\delta(\pi,S)$, which ensures that the following implication holds for $\overline{\tau}$ defined in the statement of the lemma:
\begin{align}
    C_{\hat{S}}(\pi)-\overline{C}_\delta(\pi,S) \geq \overline{\tau} \implies C_{\hat{S}}(\pi) - C_{\D'}(\pi)\geq \overline{\tau} .
\end{align}
Therefore, if $\overline{\tau}>0$ we have that 
\small
\begin{align}\nonumber
    & \underset{\hat{S}\sim\D'^n}{\mathbb{P}}[C_{\hat{S}}(\pi)-\overline{C}_\delta(\pi,S) \geq \overline{\tau} ~|~C_{\D'}(\pi) \leq C_\D(\pi)]  \\ & \leq \underset{\hat{S}\sim\D'^n}{\mathbb{P}}[C_{\hat{S}}(\pi) - C_{\D'}(\pi)\geq \overline{\tau}] \leq \exp(-2n\overline{\tau}^2)  , \nonumber
\end{align}
\normalsize
where the last upper bound follows from Hoeffding's inequality. Hence, for $\overline{\tau}>0$, using the above in \eqref{eq:bound-0} gives
\begin{align}\label{eq:bound-1}
    p_A(S') \leq \exp(-2n\overline{\tau}^2) .
\end{align}
If $\overline{\tau}=0$, the vacuous bound holds:
\begin{align}\label{eq:p-value-vacuous}
p_A(S')\leq 1 = \exp(-2n0) = \exp(-2n\overline{\tau}^2) .
\end{align}
Combining the two cases for $\overline{\tau}>0$ in \eqref{eq:bound-1} and $\overline{\tau}=0$ in \eqref{eq:p-value-vacuous} gives us the following implication:
\begin{align}
    C_{\D}(\pi)\leq \overline{C}_\delta(\pi,S) \implies p(S') \leq \exp(-2n\overline{\tau}(S)^2) . \label{eq:thm-1-implic}
\end{align}
Now, we expand the left-hand side of \eqref{eq:claim} using the definition of conditional probability:
\small
\begin{align}
    & \underset{(S,\pi)\sim(\D^m\times P)}{\mathbb{P}} [p_A(S') \leq \exp(-2n\overline{\tau}(S)^2)~|~C_{\D}(\pi)\leq \overline{C}_\delta(\pi,S)] =\\
    & \frac{\underset{(S,\pi)\sim(\D^m\times P)}{\mathbb{P}} [p_A(S') \leq \exp(-2n\overline{\tau}(S)^2)~\wedge~C_{\D}(\pi)\leq \overline{C}_\delta(\pi,S)]}{\underset{(S,\pi)\sim(\D^m\times P)}{\mathbb{P}} [C_{\D}(\pi)\leq \overline{C}_\delta(\pi,S)]} \label{eq:cond-prob}
\end{align}
\normalsize
From \eqref{eq:thm-1-implic}, we know that $\{(S,\pi)~|~ C_{\D}(\pi)\leq \overline{C}_\delta(\pi,S)\} \subseteq \{(S,\pi)~|~ p_A(S') \leq \exp(-2n\overline{\tau}(S)^2)\} $, therefore, $\underset{(S,\pi)\sim(\D^m\times P)}{\mathbb{P}} [p_A(S') \leq \exp(-2n\overline{\tau}(S)^2)~\wedge~C_{\D}(\pi)\leq \overline{C}_\delta(\pi,S)] = \underset{(S,\pi)\sim(\D^m\times P)}{\mathbb{P}} [C_{\D}(\pi)\leq \overline{C}_\delta(\pi,S)]$ which on using in \eqref{eq:cond-prob} gives the following:
\small
\begin{align}
    \underset{(S,\pi)\sim(\D^m\times P)}{\mathbb{P}} &[p_A(S') \leq \exp(-2n\overline{\tau}(S)^2)~|~C_{\D}(\pi)\leq \overline{C}_\delta(\pi,S)]\nonumber \\ & = \frac{\underset{(S,\pi)\sim(\D^m\times P)}{\mathbb{P}} [C_{\D}(\pi)\leq \overline{C}_\delta(\pi,S)]}{\underset{(S,\pi)\sim(\D^m\times P)}{\mathbb{P}} [C_{\D}(\pi)\leq \overline{C}_\delta(\pi,S)]} = 1, \label{eq:thm-1-end}
\end{align}
\normalsize
completing the proof of the claim \eqref{eq:claim} as well as the lemma.
\end{proof}

\begin{lemma}\label{lem:p-val-W}
Let the assumptions of Theorem~\ref{thm:p-value} hold. For notational simplicity and without loss of generality, we let $\delta_{W} = \delta$. Then,
\begin{align}
    \underset{(S,\pi)\sim(\D^m\times P)}{\mathbb{P}} [p_B(S') \leq \exp(-2n\underline{\tau}(S)^2)] \geq 1-\delta ,
\end{align}
where $\underline{\tau}(S):=\max\{ \underline{C}_\delta(\pi,S)-C_{S'}(\pi), 0\}$.
\end{lemma}
\begin{proof}
We prove this lemma by considering two cases: when the PAC-Bayes cost inequality in Corollary~\ref{cor:pac-bayes-lower-bound} holds, i.e., $C_\D(\pi) \geq \underline{C}_\delta(\pi,S)$, and when it does not, i.e., $C_\D(\pi) < \underline{C}_\delta(\pi,S)$; the two cases are considered in \eqref{eq:p-val-cond-begin}-\eqref{eq:p-val-cond}. In the latter case, we cannot say anything about the p-value, while in the former case, which holds with probability at least $1-\delta$, we show in \eqref{eq:claim}-\eqref{eq:thm-1-end} that $p_B(S') \leq \exp(-2n\underline{\tau}(S)^2)$.

Let us begin the proof by conditioning $\underset{(S,\pi)\sim(\D^m\times P)}{\mathbb{P}} [p_B(S') \leq \exp(-2n\overline{\tau}(S)^2)]$ as follows:
\small
\begin{align}
    \underset{(S,\pi)\sim(\D^m\times P)}{\mathbb{P}}  [&p_B(S') \leq \exp(-2n\underline{\tau}(S)^2)] = \nonumber \\ 
    \underset{(S,\pi)\sim(\D^m\times P)}{\mathbb{P}}  [&p_B(S') \leq \exp(-2n\underline{\tau}(S)^2)~|~C_{\D}(\pi)\geq \underline{C}_\delta(\pi,S)] \nonumber \\ & \underbrace{\underset{(S,\pi)\sim(\D^m\times P)}{\mathbb{P}}[C_\D(\pi) \geq \underline{C}_\delta(\pi,S)]}_{\geq 1-\delta~\text{(from~Corollary~\ref{cor:pac-bayes-lower-bound})} } + \nonumber\\
    & \hspace{-2cm} \underbrace{\underset{(S,\pi)\sim(\D^m\times P)}{\mathbb{P}} [p_B(S') \leq \exp(-2n\underline{\tau}(S)^2)~|~C_{\D}(\pi)< \underline{C}_\delta(\pi,S)]}_{\geq 0} \nonumber \\ 
    & \underbrace{\underset{(S,\pi)\sim(\D^m\times P)}{\mathbb{P}}[C_\D(\pi) < \underline{C}_\delta(\pi,S)]}_{\geq 0} \geq \label{eq:p-val-cond-begin-W} \\
    \underset{(S,\pi)\sim(\D^m\times P)}{\mathbb{P}} [&p_B(S') \leq \exp(-2n\underline{\tau}(S)^2)~| \nonumber \\ & C_{\D}(\pi)\geq \underline{C}_\delta(\pi,S)] (1-\delta) \label{eq:p-val-cond-W}
\end{align}
\normalsize
Now, we claim:
\begin{align}
    \underset{(S,\pi)\sim(\D^m\times P)}{\mathbb{P}} [&p_B(S') \leq \exp(-2n\underline{\tau}(S)^2)~| \nonumber \\ & C_{\D}(\pi)\geq \underline{C}_\delta(\pi,S)] = 1 ,\label{eq:claim-W}
\end{align}
which on using in \eqref{eq:p-val-cond-W} completes the proof of this lemma. The remainder of this proof is dedicated to establishing the claim in \eqref{eq:claim-W}.

We are given 
\begin{align}\label{eq:assump-1-W}
    C_{\D}(\pi)\geq \underline{C}_\delta(\pi,S) .
\end{align}
From Definition~\ref{def:p-value}, we have
\small
\begin{align}
    p_B(S') = \ & \underset{\hat{S}\sim\D'^n}{\mathbb{P}}[C_{\hat{S}}(\pi) \leq C_{S'}(\pi)~|~C_{\D'}(\pi) > C_\D(\pi)] \\
    = \underset{\hat{S}\sim\D'^n}{\mathbb{P}}[&\underline{C}_\delta(\pi,S) - C_{\hat{S}}(\pi) \geq \underline{C}_\delta(\pi,S) - C_{S'}(\pi)~| \nonumber \\ & C_{\D'}(\pi) > C_\D(\pi)] \\ 
    = \underset{\hat{S}\sim\D'^n}{\mathbb{P}}[&\underline{C}_\delta(\pi,S) - C_{\hat{S}}(\pi) \geq \underline{\tau}~|~C_{\D'}(\pi) > C_\D(\pi)] \label{eq:bound-0-W} .
\end{align}
\normalsize
From \eqref{eq:assump-1-W} and the assumption that the null hypothesis holds in \eqref{eq:bound-0-W}, it follows that $C_{\D'}(\pi) > \underline{C}_\delta(\pi,S)$, which ensures that the following implication holds for $\underline{\tau}$ as defined in the statement of Theorem~\ref{thm:p-value}:
\begin{align}
    \underline{C}_\delta(\pi,S) - C_{\hat{S}}(\pi) \geq \underline{\tau} \implies C_{\D'}(\pi) - C_{\hat{S}}(\pi) > \overline{\tau} .
\end{align}
Therefore, if $\underline{\tau}>0$ we have that 
\small
\begin{align}\nonumber
    \underset{\hat{S}\sim\D'^n}{\mathbb{P}}&[\underline{C}_\delta(\pi,S) - C_{\hat{S}}(\pi) \geq \underline{\tau}~|~C_{\D'}(\pi) > C_\D(\pi)] \\ & \nonumber \leq  \underset{\hat{S}\sim\D'^n}{\mathbb{P}}[C_{\D'}(\pi) - C_{\hat{S}}(\pi) > \overline{\tau}] \leq \exp(-2n\overline{\tau}^2)  ,
\end{align}
\normalsize
where the last upper bound follows from Hoeffding's inequality. Hence, for $\underline{\tau}>0$, using the above in \eqref{eq:bound-0-W} gives
\begin{align}\label{eq:bound-1-W}
    p_B(S') \leq \exp(-2n\underline{\tau}^2) .
\end{align}
If $\underline{\tau}=0$, the vacuous bound holds:
\begin{align}\label{eq:p-value-vacuous-W}
p_B(S')\leq 1 = \exp(-2n0) = \exp(-2n\underline{\tau}^2) .
\end{align}
Combining the two cases for $\overline{\tau}>0$ in \eqref{eq:bound-1-W} and $\overline{\tau}=0$ in \eqref{eq:p-value-vacuous-W} gives us the following implication:
\begin{align}
    C_{\D}(\pi) \geq \underline{C}_\delta(\pi,S) \implies p_B(S') \leq \exp(-2n\underline{\tau}(S)^2) . \label{eq:thm-1-implic2}
\end{align}
Now, we expand the left-hand side of \eqref{eq:claim-W} using the definition of conditional probability:
\small
\begin{align}
    & \underset{(S,\pi)\sim(\D^m\times P)}{\mathbb{P}} [p_B(S') \leq \exp(-2n\underline{\tau}(S)^2)~|~C_{\D}(\pi) \geq \underline{C}_\delta(\pi,S)] = \\
    & \frac{\underset{(S,\pi)\sim(\D^m\times P)}{\mathbb{P}} [p_B(S') \leq \exp(-2n\underline{\tau}(S)^2)~\wedge~C_{\D}(\pi)\geq \overline{C}_\delta(\pi,S)]}{\underset{(S,\pi)\sim(\D^m\times P)}{\mathbb{P}} [C_{\D}(\pi) \geq \underline{C}_\delta(\pi,S)]} \label{eq:cond-prob-W}
\end{align}
\normalsize
From \eqref{eq:thm-1-implic2}, we know that $\{(S,\pi)~|~ C_{\D}(\pi)\geq \underline{C}_\delta(\pi,S)\} \subseteq \{(S,\pi)~|~ p_B(S') \leq \exp(-2n\underline{\tau}(S)^2)\} $, therefore, $\underset{(S,\pi)\sim(\D^m\times P)}{\mathbb{P}} [p_B(S') \leq \exp(-2n\underline{\tau}(S)^2)~\wedge~C_{\D}(\pi)\geq \underline{C}_\delta(\pi,S)] = \underset{(S,\pi)\sim(\D^m\times P)}{\mathbb{P}} [C_{\D}(\pi)\geq \underline{C}_\delta(\pi,S)]$ which on using in \eqref{eq:cond-prob} gives the following:
\small
\begin{align}
    \underset{(S,\pi)\sim(\D^m\times P)}{\mathbb{P}} &[p_B(S') \leq \exp(-2n\underline{\tau}(S)^2)~|~C_{\D}(\pi)\geq \overline{C}_\delta(\pi,S)] \nonumber \\ & = \frac{\underset{(S,\pi)\sim(\D^m\times P)}{\mathbb{P}} [C_{\D}(\pi)\geq \underline{C}_\delta(\pi,S)]}{\underset{(S,\pi)\sim(\D^m\times P)}{\mathbb{P}} [C_{\D}(\pi)\geq \underline{C}_\delta(\pi,S)]} = 1, \label{eq:thm-1-end-W}
\end{align}
\normalsize
completing the proof of the claim \eqref{eq:claim-W} as well as the lemma.
\end{proof}

\begin{proof}[Proof of Theorem~\ref{thm:p-value}]
The proof of Theorem~\ref{thm:p-value} follows directly from Lemma~\ref{lem:p-val-O} and Lemma~\ref{lem:p-val-W}.
\end{proof}

\subsection{Proof of Lemma~\ref{lem:p-val-consistent}}
\label{app:p-val-consistent-proof}

For the readers' convenience, we restate Lemma~\ref{lem:p-val-consistent} here and provide a detailed proof.

\begin{customlem}{1}
Algorithm~\ref{alg:HT-detector} returns mutually exclusive outputs, i.e., it returns only one of the three possibilities: $\oodRightSym$, $\oodLeftSym$, or WD.
\end{customlem}
\begin{proof} 
To prove this property of the detector, we first note that for any test dataset $S'$ the detector can encounter only one of the following \emph{distinct} cases: 
\begin{enumerate}
    \item $\exp(-2n\overline{\tau}^2)\leq \alpha_A$ and $\exp(-2n\underline{\tau}^2) > \alpha_B$
    \item $\exp(-2n\overline{\tau}^2) > \alpha_A$ and $\exp(-2n\underline{\tau}^2) \leq \alpha_B$
    \item $\exp(-2n\overline{\tau}^2) > \alpha_A$ and $\exp(-2n\underline{\tau}^2) > \alpha_B$
    \item $\exp(-2n\overline{\tau}^2)\leq \alpha_A$ and $\exp(-2n\underline{\tau}^2)\leq \alpha_B$
\end{enumerate}
The first case results in $\oodRightSym$, the second case results in $\oodLeftSym$, and the third case results in WD. In the rest of this proof we show that the fourth case (i.e., upper bounds on both p-values being smaller than their respective significance levels) cannot occur.

Using Theorem~\ref{thm:derand_pacbayes_specialized} and Corollary~\ref{cor:pac-bayes-lower-bound} we can write the following:
\small
\begin{align}
    & C_{S'}(\pi) - \overline{C}_{\delta_A}(\pi,S) = C_{S'}(\pi) - C_S(\pi) - \sqrt{R_A} > 0 \\
    & \iff C_S(\pi) - C_{S'}(\pi) < -\sqrt{R_A} \\
    & \iff C_S(\pi) - C_{S'}(\pi) -\sqrt{R_B} < -\sqrt{R_A} - \sqrt{R_B} \leq 0 \\
    & \implies C_S(\pi) - C_{S'}(\pi) -\sqrt{R_B} = \underline{C}_{\delta_B}(\pi,S)-C_{S'}(\pi) < 0
\end{align}
\normalsize
Hence, we have
\begin{align}\label{eq:HT-unamb-1}
    C_{S'}(\pi) - \overline{C}_{\delta_A}(\pi,S) > 0 \implies \underline{C}_{\delta_B}(\pi,S)-C_{S'}(\pi) < 0 .
\end{align}
If $\exp(-2n\overline{\tau}^2)\leq \alpha_A < 1$ then $\overline{\tau}>0$ which further implies that $C_{S'}(\pi) - \overline{C}_{\delta_A}(\pi,S) > 0$. Hence, from \eqref{eq:HT-unamb-1} it follows that $\underline{C}_{\delta_B}(\pi,S)-C_{S'}(\pi) < 0$ which implies that $\underline{\tau} = 0$ ensuring that $\exp(-2n\underline{\tau}^2)=1>\alpha_B$. Thus, if $\exp(-2n\overline{\tau}^2)\leq \alpha_A < 1 \implies \exp(-2n\underline{\tau}^2)=1$ ensuring that case 4 cannot occur.
\end{proof}
\subsection{Proof of Theorem~\ref{thm:CI-all}}
\label{app:CI-proof}
For the readers' convenience, we restate Theorem~\ref{thm:CI-all} here and provide a detailed proof.
\begin{customthm}{\ref{thm:CI-all}}
Let $\D$ be the training distribution, $\D'$ be the test distribution, and $P$ be the posterior distribution on the space of policies obtained through the training procedure described in Section~\ref{subsec:PAC-Bayes}. Let $\delta_A,\delta'_A\in(0,1)$ such that $\delta_A+\delta'_A < 1$, $\gamma_A := \sqrt{\frac{\ln{(1/\delta'_A)}}{2n}}$, and $\Delta C_A:= C_{S'}(\pi) - \gamma_A - \overline{C}_{\delta_A}(\pi,S)$. Similarly, let $\delta_B,\delta'_B\in(0,1)$ such that $\delta_B+\delta'_B < 1$,
$\gamma_B := \sqrt{\frac{\ln{(1/\delta'_B)}}{2n}}$, and $\Delta C_B:= \underline{C}_{\delta_B}(\pi,S) - C_{S'}(\pi) - \gamma_B$. Then, 
\begin{align}
    (i) \underset{(S,\pi,S')\sim(\D^m\times P \times \D'^n)}{\mathbb{P}}&[C_{\D'}(\pi) - C_{\D}(\pi) \nonumber \\ & \geq \Delta C_A] \geq 1-\delta_A-\delta'_A , \\
    (ii) \underset{(S,\pi,S')\sim(\D^m\times P \times \D'^n)}{\mathbb{P}}&[C_{\D}(\pi) - C_{\D'}(\pi) \geq \Delta C_B ] \nonumber \\ & \geq 1-\delta_B-\delta'_B .
\end{align}
\end{customthm}
To prove Theorem~\ref{thm:CI-all}, we establish and prove Lemmas~\ref{lem:CI} and \ref{lem:CI-FN}.
\begin{lemma}\label{lem:CI}
Let the assumptions of Theorem~\ref{thm:CI-all} hold. Then, 
\begin{align} 
    \underset{(S,\pi,S')\sim(\D^m\times P \times \D'^n)}{\mathbb{P}}&[C_{\D'}(\pi) - C_{\D}(\pi) \geq \Delta C_A ]\nonumber \\ & \geq 1-\delta_A-\delta'_A .\label{eq:thm:CI}
\end{align}
\end{lemma} \vspace{-7pt}
\begin{proof}
To lower bound the difference between $C_{\D'}(\pi)$ and $C_\D(\pi)$ with high probability we obtain a lower bound on $C_{\D'}(\pi)$ which holds with probability at least $1-\delta'$ using Hoeffding's inequality in \eqref{eq:joint}-\eqref{eq:test-bound}. Then we use this bound with the PAC-Bayes bound \eqref{pointwise_pacbayes_specialized} which holds with probability at least $1-\delta$ to obtain \eqref{eq:thm:CI} by following the steps in \eqref{eq:implies}-\eqref{eq:thm-3-end}.
Let $\gamma$ be defined as in the statement of the theorem, then, using the independence of $\D'^n$ from $\D^m \times P$, we can write\footnote{Note that $C_{\D'}(\pi)$ and $C_{S'}(\pi)$ implicitly depend on $S$ because the posterior distribution $P$, from which $\pi$ is sampled, is trained on $S$.}
\begin{align}\label{eq:joint}
    \underset{(S,\pi,S')\sim(\D^m\times P \times \D'^n)}{\mathbb{P}}&[C_{\D'}(\pi) \geq C_{S'}(\pi) - \gamma] \nonumber\\
    = \int_{(S,\pi)} \underset{S'\sim \D'^n}{\mathbb{P}}&[C_{\D'}(\pi) \geq C_{S'}(\pi) - \gamma~| S,\pi]  \nonumber \\ & d(\D^m\times P)(S,\pi) .
\end{align}
For \emph{any} given $(S,\pi)$, we can apply Hoeffding's inequality to get:
\small
\begin{align} \label{eq:hoeff}
    & \underset{S'\sim \D'^n}{\mathbb{P}}[C_{\D'}(\pi) \geq C_{S'}(\pi) - \gamma]  \\ & = \underset{S'\sim\D'^n}{\mathbb{P}}[C_{S'}(\pi) - C_{\D'}(\pi)\leq \gamma] \geq 1-\exp(-2n\gamma^2) = 1-\delta' .\nonumber
\end{align}
\normalsize
Using \eqref{eq:hoeff} in \eqref{eq:joint} we get that:
\small
\begin{align}
    &\hspace{-.5cm}\underset{(S,\pi,S')\sim(\D^m\times P \times \D'^n)}{\mathbb{P}}[C_{\D'}(\pi) \geq C_{S'}(\pi) - \gamma] \nonumber \\ 
    & \geq \int_{(S,\pi)} (1-\delta') d(\D^m\times P)(S,\pi) \\
    &= (1-\delta')\int_{(S,\pi)} d(\D^m\times P)(S,\pi) = 1-\delta' \label{eq:test-bound} .
\end{align}
\normalsize
Now, observe that
\small
\begin{align}
    & C_\D(\pi) \leq \overline{C}_{\delta}(\pi,S) ~ \wedge ~ C_{\D'}(\pi) \geq C_{S'}(\pi) - \gamma \nonumber \\ & \implies C_{\D'}(\pi) - C_{\D}(\pi) \geq C_{S'}(\pi) - \gamma - \overline{C}_{\delta}(\pi,S) .\label{eq:implies}
\end{align}
\normalsize
From the implication \eqref{eq:implies}, it follows that 
\small
\begin{align}
    \underset{(S,\pi,S')\sim(\D^m\times P \times \D'^n)}{\mathbb{P}}[&C_{\D'}(\pi) - C_{\D}(\pi)~\geq \nonumber \\ & C_{S'}(\pi) - \gamma - \overline{C}_{\delta}(\pi,S)] \\
    \geq \underset{(S,\pi,S')\sim(\D^m\times P \times \D'^n)}{\mathbb{P}}[&C_\D(\pi) \leq \overline{C}_{\delta}(\pi,S) ~ \wedge \nonumber \\ & C_{\D'}(\pi) \geq C_{S'}(\pi) - \gamma] \label{eq:thm-3-int-1}
\end{align}
\normalsize
Now using the Fr\'echet inequality $\mathbb{P}[E_1 \wedge E_2] \geq \mathbb{P}[E_1] + \mathbb{P}[E_2] - 1$ (where $E_1$ and $E_2$ are arbitrary random events) on \eqref{eq:thm-3-int-1} we obtain:
\begin{align}
    \underset{(S,\pi,S')\sim(\D^m\times P \times \D'^n)}{\mathbb{P}}[&C_\D(\pi) \leq \overline{C}_{\delta}(\pi,S) ~ \wedge \nonumber \\ & C_{\D'}(\pi) \geq C_{S'}(\pi) - \gamma] \\
    \geq \underset{(S,\pi)\sim(\D^m\times P)}{\mathbb{P}}[&C_\D(\pi) \leq \overline{C}_{\delta}(\pi,S)] \ + \nonumber \\ \underset{(S,\pi,S')\sim(\D^m\times P \times \D'^n)}{\mathbb{P}}[&C_{\D'}(\pi) \geq C_{S'}(\pi) - \gamma] - 1 \label{eq:thm-3-int-2}\\
    \geq 1-\delta - \delta' ,\hspace{0.3cm}& \label{eq:thm-3-end}
\end{align}
where the last inequality follows by using \eqref{pointwise_pacbayes_specialized} and \eqref{eq:test-bound} in \eqref{eq:thm-3-int-2}. Finally, using \eqref{eq:thm-3-end} in \eqref{eq:thm-3-int-1} completes the proof.
\end{proof} 
\begin{lemma}\label{lem:CI-FN}
Let the assumptions of Theorem~\ref{thm:CI-all} hold. Then,
\begin{align}
    \underset{(S,\pi,S')\sim(\D^m\times P \times \D'^n)}{\mathbb{P}}&[C_{\D}(\pi) - C_{\D'}(\pi) \geq \Delta C_B ] \nonumber \\ & \geq 1-\delta_B-\delta'_B .
\end{align}
\end{lemma}
\begin{proof}
Let $\gamma$ be defined as in the statement of the theorem, then, using the independence of $\D'^n$ from $\D^m \times P$, we can write\footnote{Note that $C_{\D'}(\pi)$ and $C_{S'}(\pi)$ implicitly depend on $S$ because the posterior distribution $P$, from which $\pi$ is sampled, is trained on $S$.}
\begin{align}\label{eq:joint_fn}
    \underset{(S,\pi,S')\sim(\D^m\times P \times \D'^n)}{\mathbb{P}}&[C_{\D'}(\pi) \leq C_{S'}(\pi) + \gamma] \nonumber\\
    = \int_{(S,\pi)} \underset{S'\sim \D'^n}{\mathbb{P}}&[C_{\D'}(\pi) \leq C_{S'}(\pi) + \gamma~|~S,\pi] \nonumber \\ & d(\D^m\times P)(S,\pi) .
\end{align}
For \emph{any} given $(S,\pi)$, we can apply Hoeffding's inequality to get:
\small
\begin{align}\label{eq:hoeff_fn}
    & \underset{S'\sim \D'^n}{\mathbb{P}}[C_{\D'}(\pi) \leq C_{S'}(\pi) + \gamma] \\ & = \underset{S'\sim\D'^n}{\mathbb{P}}[C_{\D'}(\pi) - C_{S'}(\pi)\leq \gamma] \geq 1-\exp(-2n\gamma^2) = 1-\delta' . \nonumber 
\end{align}
\normalsize
Using \eqref{eq:hoeff_fn} in \eqref{eq:joint_fn} we get that:
\small
\begin{align}
    &\hspace{-0.5cm} \underset{(S,\pi,S')\sim(\D^m\times P \times \D'^n)}{\mathbb{P}}[C_{\D'}(\pi) \leq C_{S'}(\pi) + \gamma] \nonumber \\ & \geq \int_{(S,\pi)} (1-\delta') d(\D^m\times P)(S,\pi) \\
    & = (1-\delta')\int_{(S,\pi)} d(\D^m\times P)(S,\pi) = 1-\delta' \label{eq:test-bound2} .
\end{align}
\normalsize
Now, observe that
\small
\begin{align}
    & C_\D(\pi) \geq \underline{C}_\delta(\pi,S) ~ \wedge ~ C_{\D'}(\pi) \leq C_{S'}(\pi) + \gamma \nonumber \\ & \implies C_{\D}(\pi) - C_{\D'}(\pi) \geq \underline{C}_\delta(\pi,S) - C_{S'}(\pi) - \gamma .\label{eq:implies_fn}
\end{align}
\normalsize
Hence, 
\small
\begin{align}
    \underset{(S,\pi,S')\sim(\D^m\times P \times \D'^n)}{\mathbb{P}}[&C_{\D}(\pi) - C_{\D'}(\pi) \nonumber \\ & \geq \underline{C}_\delta(\pi,S) - C_{S'}(\pi) - \gamma] \\
    \geq \underset{(S,\pi,S')\sim(\D^m\times P \times \D'^n)}{\mathbb{P}}[&C_\D(\pi) \geq \underline{C}_\delta(\pi,S) ~ \wedge \nonumber \\ & C_{\D'}(\pi) \leq C_{S'}(\pi) + \gamma] \\
    \geq \underset{(S,\pi)\sim(\D^m\times P)}{\mathbb{P}}[C_\D(\pi) &\geq \underline{C}_\delta(\pi,S)] \ + \nonumber \\ \underset{(S,\pi,S')\sim(\D^m\times P \times \D'^n)}{\mathbb{P}}[&C_{\D'}(\pi) \leq C_{S'}(\pi) + \gamma] - 1 \\
    \geq 1-\delta - \delta' \hspace{1.5cm}& ,
\end{align}
\normalsize
where the first inequality follows from \eqref{eq:implies_fn}, the second inequality follows from Fr\'echet inequalities, and the third inequality follows from Theorem~\ref{thm:derand_pacbayes_specialized} and \eqref{eq:test-bound2}.
\end{proof}
\begin{proof}[Proof of Theorem~\ref{thm:CI-all}]
The proof of Theorem~\ref{thm:CI-all} follows directly from Lemma~\ref{lem:CI} and Lemma~\ref{lem:CI-FN}.
\end{proof}
\subsection{Proof of Lemma~\ref{lem:CI-unambiguous}}
\label{app:CI-consistent-proof}

For the readers' convenience, we restate Lemma~\ref{lem:CI-unambiguous} here and provide a detailed proof.

\begin{customlem}{\ref{lem:CI-unambiguous}}
Algorithm~\ref{alg:CI-detector} returns mutually exclusive outputs, i.e., it returns one of three possibilities: $\oodRightSym$, $\oodLeftSym$, or WD.
\end{customlem}
\begin{proof} 
To prove this property, we note that the detector can encounter only one of the following four distinct cases involving $\Delta C_A$ and $\Delta C_B$: 
\begin{enumerate}
    \item $\Delta C_A > 0$ and $\Delta C_B < 0$
    \item $\Delta C_A \leq 0$ and $\Delta C_B \geq 0$
    \item $\Delta C_A \leq 0$ and $\Delta C_B < 0$
    \item $\Delta C_A > 0$ and $\Delta C_B \geq 0$
\end{enumerate}
The first case results in OOD-adverse, the second case results in OOD-benign, and the third case results in WD.  In the rest of this proof we show that the fourth case cannot occur. From the definition of $\Delta C_A$ we have that:
\small
\begin{align}
    \Delta C_A & > 0  \iff C_{S'}(\pi) - \gamma_A - \overline{C}_{\delta_A}(\pi,S) > 0 \\
    \iff&\ C_{S'}(\pi) - \gamma_A - C_S({\pi}) - \sqrt{R_A} > 0 \\
    \iff&\ C_{S}(\pi) - C_{S'}({\pi}) <  -\gamma_A -\sqrt{R_A} \\
    \iff&\ C_{S}(\pi) - C_{S'}({\pi}) - \gamma_B - \sqrt{R_B}  \nonumber \\ & < -\gamma_A -\sqrt{R_A} - \gamma_B - \sqrt{R_B} \\
    \iff&\ \underline{C}_{\delta_B}(\pi,S) - C_{S'}(\pi) - \gamma_B \nonumber \\ &  < -\gamma_A -\sqrt{R_A} - \gamma_B - \sqrt{R_B} < 0 \\
    \implies&\ \Delta C_B < 0
\end{align}
\normalsize
We have shown that $\Delta C_A > 0 \implies \Delta C_B < 0$. Thus case 4 is not possible and the proof is complete.
\end{proof}
\subsection{Training with Backpropogation}
\label{ap:backprop alg}

In this section, we describe a method to minimize the upper bound in Theorem~\ref{thm:derand_pacbayes_specialized} using backpropogation. We make use of multivariate Gaussian distributions $\N_\psi$ with diagonal covariance $\Sigma_s := \text{diag}(s)$ where $\psi:=(\mu,\log s)$. When training the posterior distribution $P$, we would like to take gradient steps directly with respect to $\psi$. However, this would require backpropagation through $\EE_{w \sim \N_\psi}C_E(\pi_w)$. We follow a similar procedure as in \cite{Dziugiate17} and achieve the desired result of minimizing the upper bound in Inequality~\eqref{pointwise_pacbayes_specialized} using an unbiased estimate of $\EE_{w \sim \N_\psi}C_E(\pi_w)$:
\begin{equation}
    \frac{1}{k}\sum_{i=1}^k C_E(\pi_{w_i}), \ \ \ \ w_i \sim \N_\psi \ \forall \ i \in \{1,2, \dots, k\} .
\end{equation} 
The resulting approach is presented in Algorithm~\ref{alg:pacbayes backprop}. Note that the algorithm must be deterministic in order to maintain the assumptions of Theorem~\ref{thm:derand_pacbayes_specialized}. We achieve this by training with a fixed seed for generating random numbers. Additionally, note that the backpropagation requires a gradient taken through $D_2(\N_\psi \| \N_{\psi_0})$. We make use of the analytical form for the R\'enyi divergence between two multivariate Gaussian distributions, presented in \cite{Gil13}, in order to tractably compute the gradients. 
\begin{align}
D_2(\N_\psi \| \N_{\psi_0}) & = D_2\big(\N(\mu, \Sigma_s) \| \N(\mu_0, \Sigma_{s_0})\big) \\ & = (\mu - \mu_0)^T\Sigma_2(\mu - \mu_0) - \frac{1}{2}\ln\frac{| \Sigma_2||\Sigma_s|}{|\Sigma_{s_0}|^{2}}, \nonumber
\end{align}
where $\Sigma_{2} = 2\Sigma_{s_0} - \Sigma_{s}$. We also note that there is a restriction on how far the posterior's variance can drift from the prior. The following expression must be satisfied for $D_2(\N_\psi \| \N_{\psi_0})$ to be finite \citep{Gil13}:
\begin{equation}
    2\Sigma_s^{-1} - \Sigma_{s_0}^{-1} \succ 0.
\end{equation}
In practice, we project any problematic variances into the range of allowable variances. 

\begin{algorithm}[H]
    \caption{PAC-Bayes Bound Minimization via Backpropagation}
    \label{alg:pacbayes backprop}
\begin{algorithmic}
    \State \textbf{Input}: Fixed prior distribution $\N_{\psi_0}$ over policies, fixed seed for random number generation
    \State \textbf{Input}: Training dataset $S$, learning rate $\gamma$ 
    \State \textbf{Output}: Optimized $\psi^*$
    \While{not converged}
    \State Sample $w_i \sim \N_\psi \ \forall \ i \in \{1, 2, ..., k\}$ 
    \State $B \leftarrow \frac{1}{mk} \sum_{E\in S} \sum_{i=1}^k C_E(\pi_{w_i}) + \sqrt{R}$ 
    \State $\psi \leftarrow \psi - \gamma \nabla_{\psi} B$ 
    \EndWhile
\end{algorithmic}
\end{algorithm}

After training, since we have used the pointwise PAC-Bayes bound in Theorem~\ref{thm:derand_pacbayes_specialized}, we compute the upper bound with a single $w \sim \N_\psi$ in contrast to traditional PAC-Bayes bounds. Thus, the resulting policy $\pi_w$ is deterministic and applicable in a broad range of settings, including ones which require a pre-trained network. The resulting policy carries a PAC-Bayes guarantee.

\subsection{Training with Evolutionary Strategies}
\label{app:reg-NES}

To train robot control policies in settings where backpropagation is not feasible (e.g. presence of a ``blackbox" in the form of a simulator or robot hardware in the forward pass), we use Evolutionary Strategies (ES) which is a class of blackbox optimizers \cite{Wierstra14}. ES addresses this challenge by estimating the gradient via a Monte-Carlo estimator:
\begin{align}
\nabla_\psi C_S(\N_\psi) & \coloneqq \frac{1}{m} \sum_{E \in S} \nabla_\psi\underset{w \sim \N_\psi}{\EE} [C_E(\pi_w)] \nonumber \\ & = \frac{1}{m} \sum_{E \in S}\underset{w \sim \N_\psi}{\EE}[C_E(\pi_w)\nabla_\psi \ln \N_{\psi}(w)] . \label{eq:emp-grad-nes}
\end{align}
Although we can compute the gradient of the regularizer analytically (as mentioned in Appendix~\ref{ap:backprop alg}), using different methods to estimate the gradient of the empirical cost (ES) and the gradient of the regularizer (analytically) results in poor convergence. To alleviate this, we estimate the regularizer's gradient using ES as well by leveraging the expectation form of R\'enyi divergence in Theorem~\ref{thm:derand_pacbayes_specialized}. This takes the following form: 
\small
\begin{align}\label{eq:PAC-Bayes-NES}
    \nabla_\psi&(C_S(\N_\psi) + \sqrt{R}) = \frac{1}{m} \sum_{E \in S}\underset{w \sim \N_\psi}{\EE} \\
    & \Bigg[ \underbrace{\bigg( C_E(\pi_w) + \frac{\mathrm{e}^{\ln\big(\frac{\N_\psi(w)}{\N_{\psi_0}(w)}\big) - D_2(\N_\psi||\N_{\psi_0})}}{4m\sqrt{R}} \bigg)\nabla_\psi \ln \N_{\psi}(w)}_{\tilde{C}_E(w)} \Bigg] . \nonumber
\end{align}
\normalsize
\subsubsection{Derivation of \eqref{eq:PAC-Bayes-NES}}
To derive \eqref{eq:PAC-Bayes-NES}, note that
\begin{align}
    \nabla_\psi(C_S(\N_\psi) + \sqrt{R}) & = \nabla_\psi C_S(\N_\psi) + \nabla_\psi\sqrt{R}\nonumber \\ & = \nabla_\psi C_S(\N_\psi) + \frac{1}{2\sqrt{R}}\nabla_\psi R  . \label{eq:PAC-Bayes-ES-1}
\end{align}
From \eqref{eq:emp-grad-nes} we know the gradient for $\nabla_\psi C_S(\N_\psi)$. In the rest of this derivation, therefore, we will focus on the computing the gradient of the second term. 

Note that the R\'enyi divergence for multivariate Gaussian distributions can be written as:
\begin{align}\label{eq:renyi-div-2-normal}
    D_2(\N_\psi||\N_{\psi_0}) = \ln \Bigg(\underset{w \sim \N_{\psi_0}}{\EE}\bigg[ \bigg( \frac{\N_\psi(w)}{\N_{\psi_0}(w)} \bigg)^2 \bigg]\Bigg).
\end{align}
Let 
\begin{align}
    \eta := \underset{w\sim\N_{\psi_0}}{\EE} \bigg[ \bigg( \frac{\N_\psi(w)}{\N_{\psi_0}(w)} \bigg)^2 \bigg] ,
\end{align}
then, using \eqref{eq:renyi-div-2-normal} we have that 
\begin{align}\label{eq:D2-eta}
D_2(\N_\psi||\N_{\psi_0}) = \ln \eta
\end{align}
which allows us to express $R$ as
\begin{align}
    R = \frac{\ln \eta + \ln (2\sqrt{m}/(\delta/2)^3)}{2m} .
\end{align}
Hence,
\begin{align}\label{eq:R-grad}
    \nabla_\psi R = \frac{1}{2m\eta}\nabla_\psi \eta = \frac{\mathrm{e}^{-D_2(\N_\psi||\N_{\psi_0})}}{2m}\nabla_\psi \eta ,
\end{align}
where the last equality follows from \eqref{eq:D2-eta}.

For computing the gradient using ES, we require the cost to be an expectation over the posterior, however, $\eta$ is an expectation on the prior. To address this we perform a change of measure which gives us the following:
\begin{align}\label{eq:change-of-measure}
    \eta = \underset{w\sim\N_{\psi_0}}{\EE} \bigg[ \bigg( \frac{\N_\psi(w)}{\N_{\psi_0}(w)} \bigg)^2 \Bigg] = \underset{w\sim\N_\psi}{\EE} \bigg[ \frac{\N_\psi(w)}{\N_{\psi_0}(w)} \Bigg] .
\end{align}
Using \eqref{eq:change-of-measure} in \eqref{eq:R-grad} gives us
\begin{align}
    \nabla_\psi R & = \frac{\mathrm{e}^{-D_2(\N_\psi||\N_{\psi_0})}}{2m}\nabla_\psi \underset{w\sim\N_\psi}{\EE} \bigg[ \frac{\N_\psi(w)}{\N_{\psi_0}(w)} \Bigg] \label{eq:R-ES-grad} \\ & = \frac{\mathrm{e}^{-D_2(\N_\psi||\N_{\psi_0})}}{2m} \underset{w\sim\N_\psi}{\EE} \bigg[ \frac{\N_\psi(w)}{\N_{\psi_0}(w)} \nabla_\psi \ln \N_\psi(w) \Bigg] . \nonumber
\end{align}
Using \eqref{eq:R-ES-grad} and \eqref{eq:emp-grad-nes} in \eqref{eq:PAC-Bayes-ES-1} and combining the expectation terms gives
\small
\begin{align}
    \nabla_\psi(&C_S(\N_\psi) + \sqrt{R}) = \frac{1}{m}\sum_{E \in S}\underset{w \sim \N_\psi}{\EE} \\
    & \Bigg[ \bigg( C_E(\pi_w) + \frac{\mathrm{e}^{ - D_2(\N_\psi||\N_{\psi_0})}}{4m\sqrt{R}} \frac{\N_\psi(w)}{\N_{\psi_0}(w)} \bigg)\nabla_\psi \ln \N_{\psi}(w) \Bigg].\nonumber
\end{align}
\normalsize
Finally, we note that the dimensionality $d$ of $w$ can be large, in which case the term $\N_\psi(w)/\N_{\psi_0}(w)$ is numerically unstable because it involves the product of $d$ terms. Hence, we express $\N_\psi(w)/\N_{\psi_0}(w)$ as $\mathrm{e}^{\ln( \N_\psi(w)/\N_{\psi_0}(w))}$ which gives us \eqref{eq:PAC-Bayes-NES} as the final form of the gradient.

\subsubsection{Training algorithm}
The gradient of the PAC-Bayes upper bound is estimated from \eqref{eq:PAC-Bayes-NES}. Since Theorem~\ref{thm:derand_pacbayes_specialized} requires the training algorithm to be deterministic, we train with a fixed seed. The psuedo-code for our training is provided in Algorithm~\ref{alg:pacbayes ES}. After training, a single $w$ is drawn from $\N_{\psi^*}$, which corresponds to a policy $\pi_w$, and the derandomized PAC-Bayes bound is computed for this policy.
\begin{algorithm}[h]
    \caption{PAC-Bayes Bound Minimization via ES}
    \label{alg:pacbayes ES}
\begin{algorithmic}
    \State \textbf{Input}: Fixed prior distribution $\N_{\psi_0}$ over policies, fixed seed for random number generation
    \State \textbf{Input}: Training dataset $S$, learning rate $\gamma$
    \State \textbf{Output}: Optimized $\psi^*$
    \While{not converged}
    \State Sample $w_i \sim \N_\psi \ \forall \ i \in \{1, 2, ..., k\}$ 
    \State $\texttt{grad} \leftarrow \frac{1}{mk} \sum_{E\in S} \sum_{i=1}^k \tilde{C}_E(w_i)$
    \State $\psi \leftarrow \psi - \gamma \cdot \texttt{grad}$ 
    \EndWhile
\end{algorithmic}
\end{algorithm}

\subsection{Additional Experimental Details and Results}

\begin{figure}[t]
\centering
\includegraphics[width=0.9\columnwidth]{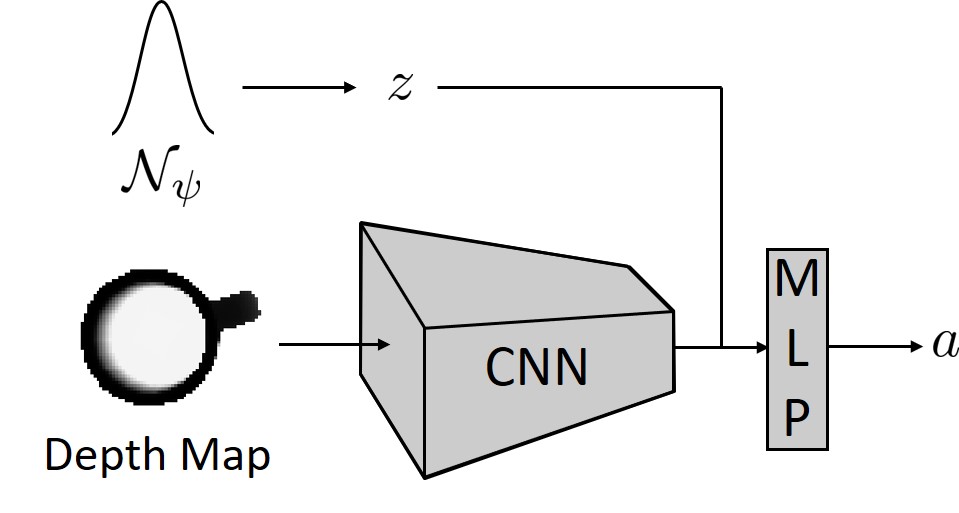}
\caption{Network architecture for the manipulator's grasping control policy. }
\label{fig:grasp-network}
\end{figure}

\subsubsection{Robotic grasping}
\label{app:manipulator-exp}

\textbf{Training platform.} Training was performed on a \texttt{Lambda Blade} server with \texttt{2x Intel Xeon Gold 5220R} (96 CPU threads) and 768 GB RAM.

\textbf{Distributions on the initial position of the mugs.} 
For all datasets, mugs are placed upright on the table with random yaw orientations sampled from the uniform distribution $\mathcal{U}([-\pi~\mathrm{rad}, \pi~\mathrm{rad}])$. 
The distributions listed on the mug's placement were used to generate the plot in Figure~\ref{fig:mug-pos} and in Figure~\ref{fig:mug-pos-hw}. Note that distribution (6) enumerated below was not used for the hardware experiments as it was outside the visual field of the camera mounted (for the hardware setup see Figure~\ref{fig:franka-panda-hw}). 
\begin{enumerate}
    \item $\mathcal{U}([0.45~\mathrm{m}, 0.55~\mathrm{m}]\times[-0.05~\mathrm{m},0.05~\mathrm{m}])$ (training distribution)
    \item $\mathcal{U}([0.40~\mathrm{m}, 0.60~\mathrm{m}]\times[-0.10~\mathrm{m},0.10~\mathrm{m}])$
    \item $\mathcal{U}([0.35~\mathrm{m}, 0.65~\mathrm{m}]\times[-0.15~\mathrm{m},0.15~\mathrm{m}])$
    \item $\mathcal{U}([0.30~\mathrm{m}, 0.70~\mathrm{m}]\times[-0.20~\mathrm{m},0.20~\mathrm{m}])$
    \item $\mathcal{U}([0.25~\mathrm{m}, 0.75~\mathrm{m}]\times[-0.25~\mathrm{m},0.25~\mathrm{m}])$
    \item $\mathcal{U}([0.20~\mathrm{m}, 0.80~\mathrm{m}]\times[-0.30~\mathrm{m},0.30~\mathrm{m}])$
\end{enumerate}

\textbf{Control policy architecture.} The control policy architecture is shown in Figure~\ref{fig:grasp-network}. The weights of the DNN are from \cite{Ren20} and the training is warm-started with the posterior in \citep[Appendix~A5.1]{Ren20}. 

\subsubsection{Vision-based obstacle avoidance with a drone}
\label{app:navigation-exp}

The approximate $C_\D(\pi)$ (estimated with $50,000$ held-out environments) is $0.149$; PAC-Bayes thus provides a strong bound.

\textbf{Environment generation.} Training environments have $9$ obstacles and have at least one gap which is wide enough to navigate through. We generate environments by randomly placing a set of cylindrical obstacles whose locations are sampled from the uniform distribution $\mathcal{U}([4.5~\mathrm{m}, 7~\mathrm{m}]\times[-3.5~\mathrm{m}, 3.5~\mathrm{m}])$ relative to the drone's starting point.

\textbf{Training the prior.} Training takes place completely in simulation. To allow for accurate sim-to-real transfer, the motion primitives are recorded trajectories of open-loop control inputs for the Parrot Swing hardware platform. We record multiple rollouts of each open-loop control policy. In simulation, when the policy selects a motion primitive, we randomly select one of the corresponding recorded trajectories to run. 
We train the prior $\N_{\psi_0}$ over policies by transforming the problem into a supervised learning setting. For each of $10{,}000$ training environment in $S$ the policy receives a depth map. Leveraging the simulation, we simulate each primitive (sampled uniformly from the set of recorded trajectories for that primitive) through each environment. We generate a label for each depth map by recording the minimum distance to an obstacle achieved by each of the primitives and passing the vector of distances through a softmax transformation. Note that even in simulation, we do not assume knowledge of the exact location of obstacles and record the closest distance as viewed by the robot's 120$^\circ$ field of view depth sensor. These depth maps and softmax labels can then be used for training the prior over policies in a supervised learning setting. We use the cross-entropy loss to train . The result is a policy trained to assign larger values to motion primitives which achieve a larger distance from obstacles.

\textbf{Training platform.} Training was performed on a desktop computer with an \texttt{Intel i7-8700k CPU} (12 CPU threads) and an \texttt{NVIDIA Titan Xp GPU} with 32 GB RAM.

\begin{figure}[h]
\vspace{2mm}
\begin{center}
\includegraphics[width=0.9\columnwidth]{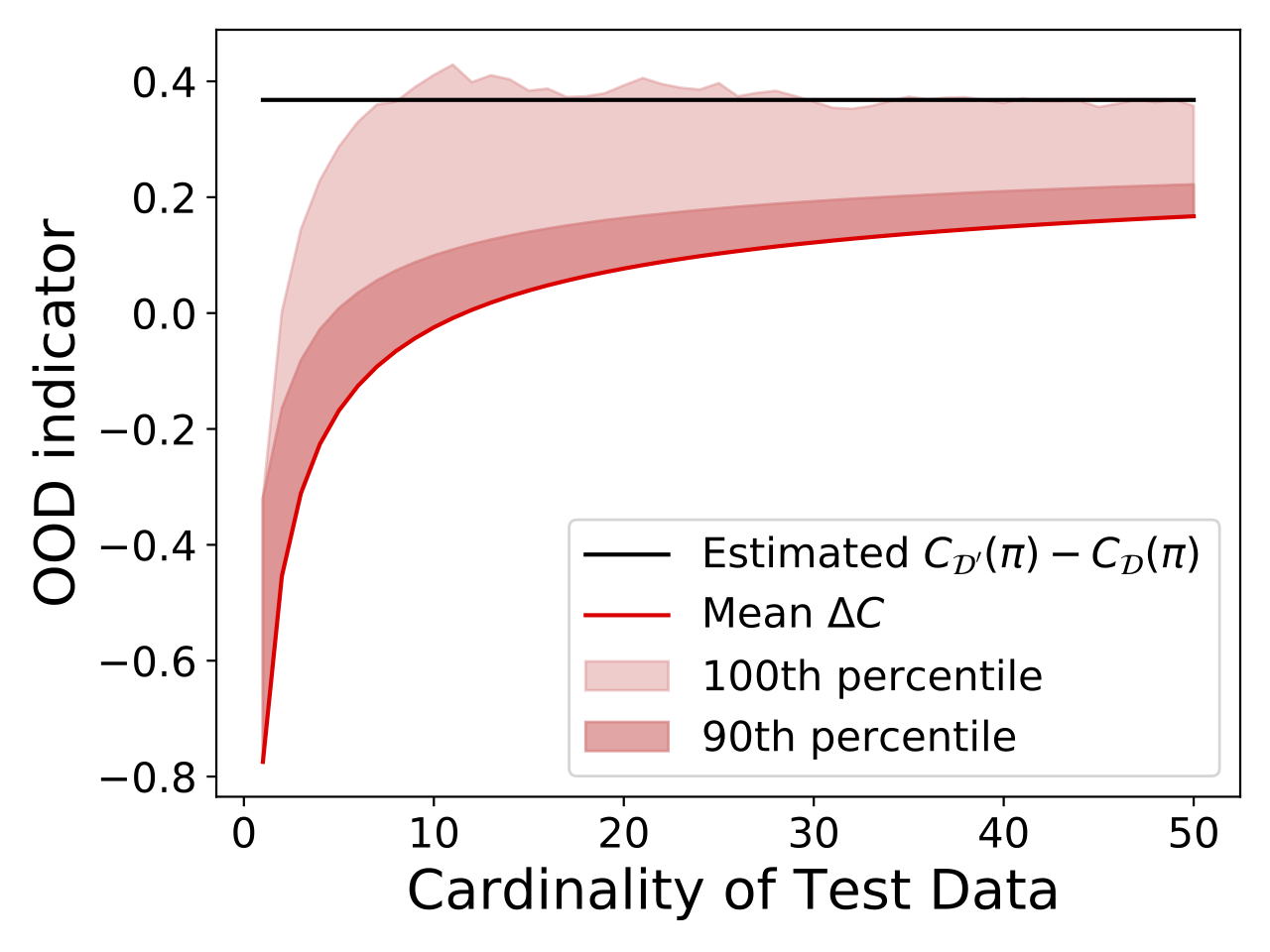}
\caption{\label{fig:swingsim_lowerbound} Numerical validation of lower bound in Theorem~\ref{thm:CI-all}.}
\end{center}
\end{figure}

\textbf{Numerical validation of Theorem~\ref{thm:CI-all}} We numerically validate our confidence bound in Figure~\ref{fig:swingsim_lowerbound}. We plot (i) the difference $C_{\mathcal{D}'}(\pi)$ - $C_{\mathcal{D}}(\pi)$ (estimated via exhaustive sampling of environments), (ii) the maximum computed lower-bound on $C_{\mathcal{D}'}(\pi)$ - $C_{\mathcal{D}}(\pi)$ (computed using a confidence level of 0.9) over $500{,}000$ datasets $S'$, and (iii) the $90\mathrm{th}$ percentile value of the bound over the $100{,}000$ datasets. As guaranteed by Theorem~\ref{thm:CI-all}, the bound is valid greater than $90\%$ of the time. 

\end{document}